\documentclass[accepted,specialissue]{melba}

\usepackage{mwe} 

\usepackage{amsmath,amsfonts}
\usepackage{hyperref}
\usepackage{orcidlink}
\usepackage{cite}
\usepackage{algorithmic}
\usepackage{subcaption}
\usepackage{graphicx}
\usepackage{textcomp}
\usepackage{multirow}
\usepackage{adjustbox}
\usepackage{booktabs}
\usepackage{hhline}
\usepackage{makecell}
\definecolor{mycolor}{rgb}{0, 0, 0}
\definecolor{revisecolor}{rgb}{0, 0, 0}

\melbaid{2025:015}  
\doi{10.59275/j.melba.2025-de23}
\melbaauthors{Yan, Chen, Xue, Du, Vilouras, Tsaftaris, and McDonagh}  
\email{Junyu.Yan@ed.ac.uk}
\volume{3}
\firstpageno{347}  
\melbayear{2025}  
\datesubmitted{2025-03-31}  
\datepublished{2025-07-31}  

\melbaspecialissue{Special issue on Fairness of AI in Medical Imaging (FAIMI)}
\melbaspecialissueeditors{Veronika Cheplygina, Aasa Feragen, Andrew King, Ben Glocker, Enzo Ferrante, Eike Petersen, Esther Puyol-Antón, Melanie Ganz-Benjaminsen}

\ShortHeadings{SWiFT: Soft-Mask Weight Fine-tuning for Bias Mitigation}{Yan, Chen, Xue, Du, Vilouras, Tsaftaris, and McDonagh}

\title{SWiFT: Soft-Mask Weight Fine-tuning for Bias Mitigation}

\author{
\firstname Junyu \surname Yan\aff{1}\orcid{0009-0003-0311-5559},
\firstname Feng \surname Chen\aff{1}\orcid{0000-0003-2915-599X},
\firstname Yuyang \surname Xue\aff{1}\orcid{0000-0002-2281-9418},  
\firstname Yuning \surname Du\aff{1}\orcid{0009-0007-4995-5472}, 
\firstname Konstantinos \surname Vilouras\aff{1}\orcid{0009-0003-5910-9748}, 
\firstname Sotirios A. \surname Tsaftaris\aff{1}\orcid{0000-0002-8795-9294},
\firstname Steven \surname McDonagh\aff{1}\orcid{0000-0001-7025-5197}
}

\affiliations{
	\num 1 \addr School of Engineering, University of Edinburgh, EH9 3FB Edinburgh, UK \\
}

\abstract{Recent studies have shown that Machine Learning (ML) models can exhibit bias in real-world scenarios, posing significant challenges in ethically sensitive
domains such as healthcare. Such bias can negatively affect model fairness, model generalization abilities and further risks amplifying social discrimination. 
There is a need to remove biases from trained models.
Existing debiasing approaches often necessitate access to original training data and need extensive model retraining; they also typically exhibit trade-offs between model fairness and discriminative performance. 
To address these challenges, we propose Soft-Mask Weight Fine-Tuning (SWiFT), a debiasing framework that efficiently improves fairness while preserving discriminative performance with much less debiasing costs. 
Notably, SWiFT requires only a small external dataset and only a few epochs of model fine-tuning.
The idea behind SWiFT is to first find the relative, and yet distinct, contributions of model parameters to both bias and predictive performance. 
Then, a two-step fine-tuning process updates each parameter with different gradient flows defined by its contribution. 
Extensive experiments with three bias sensitive attributes (gender, skin tone, and age) across four dermatological and two chest X-ray datasets demonstrate that SWiFT can consistently reduce model bias while achieving competitive or even superior diagnostic accuracy under common fairness and accuracy metrics, compared to the state-of-the-art. Specifically, we demonstrate improved model generalization ability as evidenced by superior performance on several out-of-distribution (OOD) datasets.
Our code is available at: \url{https://github.com/vios-s/SWiFT}.}


\begin{document}

\twocolumn[\maketitle]

\section{Introduction}
\label{sec:introduction}

\begin{figure}[t]
    \centering
    \includegraphics[width=0.9\linewidth]{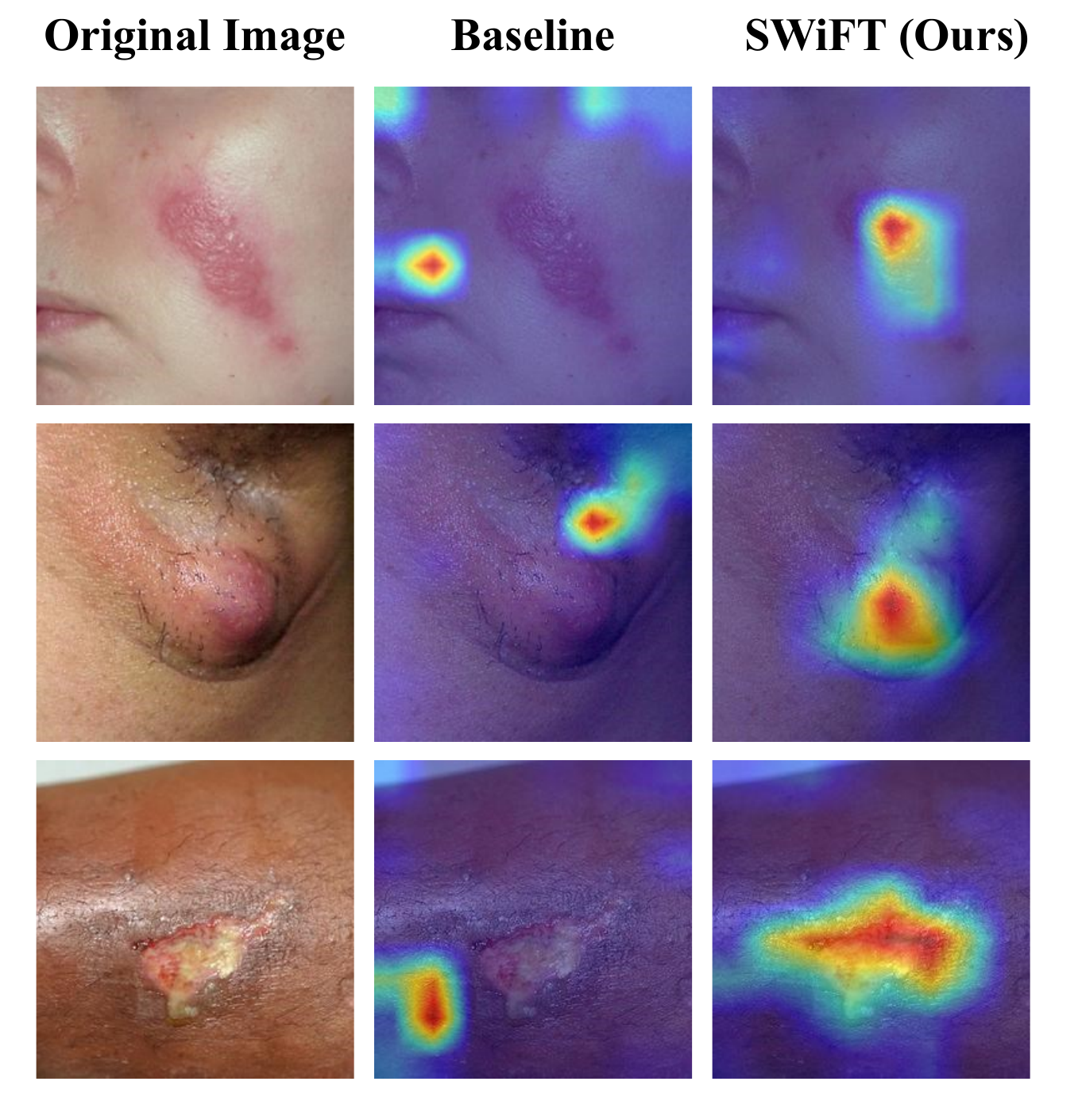}
    \caption{\textcolor{revisecolor}{An illustration of melanoma classification bias using the ISIC dataset. \textbf{Left}: column shows original images. \textbf{Middle}: Class Activation Maps (CAM) generated by a standard ERM pre-trained model (Baseline). \textbf{Right}: CAMs after debiasing with our method, SWiFT. High and low activation values are indicated with red and blue, respectively. Rows correspond to different skin tones: light skin (top), dark skin (middle and bottom). The baseline ERM classifier heavily relies on bias features (i.e., skin tone) and performs poorly on images with dark skin. In contrast, SWiFT primarily captures disease-relevant features and performs better on images with different skin tones.}}
    \label{fig.taskdescrip}
\end{figure}
\enluminure{M}achine Learning (ML) has revolutionized medical imaging applications, driving breakthroughs in areas such as computer-aided diagnosis, disease progression monitoring, and radiotherapy planning \citep{mehrabi2021survey}. 
Despite these successes, a critical concern has emerged: ML models often exhibit bias toward certain sub-populations defined by sensitive human attributes such as age, skin tone and gender. 
This bias is prevalent across various medical image analysis models, regardless of the training data modalities (e.g.~dermatological images~\citep{bevan2022skin}, X-rays~\citep{seyyed2020chexclusion}, MRI~\citep{puyol2021fairness}) or the body parts involved (e.g.~skin~\citep{bissoto2020debiasing}, chest~\citep{marcinkevics2022debiasing}).
For example, 
\citet{bevan2022skin} demonstrated that a melanoma diagnosis system trained predominantly on light-skinned images performs poorly on patients with dark skin tones. In this case, the model does not learn the correct classification strategy based on skin lesions, but rather shows a preference for erroneous correlations between sensitive attributes (i.e.~skin tone) and diagnosis, as shown in Figure~\ref{fig.taskdescrip}. 
\textcolor{mycolor}{Such biased decision-making systems can lead to disparate diagnostic performance across demographic groups, poor generalization to scenarios where these spurious correlations are missing, and further perpetuate and exacerbate social discrimination~\citep{xu2023fairness}. Thus, there is an urgent need to investigate bias mitigation strategies to improve fairness, out-of-distribution generalization ability, and trustworthiness of AI in healthcare.}

Several methods have been proposed to mitigate bias and improve fairness and generalizability in medical imaging analysis; however, two critical challenges remain. Firstly, most existing debiasing approaches focus on modifying datasets before training (i.e.~pre-processing~\citep{puyol2021fairness}) or incorporating fairness during training (i.e.~in-processing~\citep{bissoto2020debiasing}). Although these techniques can produce fair predictions with comparable discriminative performance across subgroups, they \textbf{\textit{require access to the original training data and involve computationally intensive model retraining}.} This dependency on large-scale datasets and the need for retraining limit their scalability in real-world scenarios. Yet,  
recent work~\citep{kirichenkolast} suggests that even when neural networks exhibit significant bias, they still learn features, necessary for strong classification performance, sufficiently well. This insight indicates that debiasing does not necessarily require learning from scratch; instead, it can be achieved through post-processing techniques such as fine-tuning pre-trained models. Moreover, leveraging the knowledge embedded in pre-trained models allows for efficient optimization and fast convergence~\citep{hendrycks2019using}. Therefore, we conjecture that debiasing is achievable with minimal fine-tuning on a small, task-specific dataset.

\begin{figure*}[t]
    \centering
    \includegraphics[width=1\linewidth,keepaspectratio]{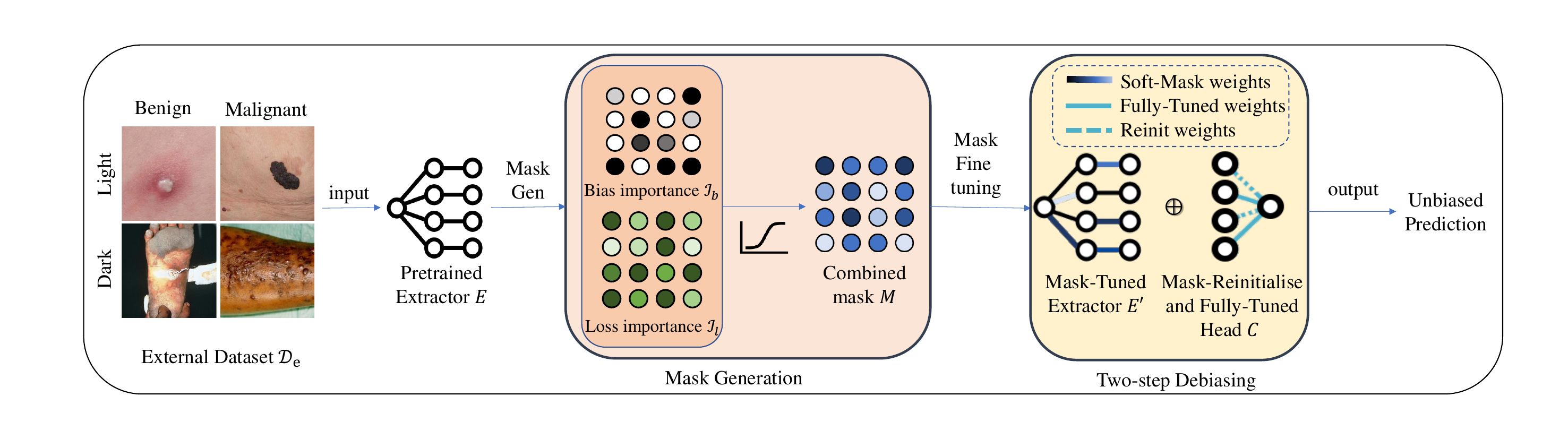}
    \caption{SWiFT is a masked-based fine-tuning post-processing approach. (1) A soft-mask is generated by calculating weight importance to the bias and prediction functions. (2) The parameters of the feature extractor $E(\cdot)$ are selectively fine-tuned using the soft-mask to debias features, followed by fine-tuning a masked re-initialized classification head $C(\cdot)$ to re-combine debiased features.}\label{fig:method}
\end{figure*} 

The second major challenge in debiasing is the \textbf{\textit{trade-off between mitigating bias and preserving the model's discriminative ability}}. 
A biased model trained with standard empirical risk minimization (ERM) often encodes two types of features: \emph{core} features that should causally contribute to prediction and \emph{bias} features that likely capture spurious correlations or irrelevant information. These features are often entangled in complex ways~\citep{le2023last}. Existing debiasing techniques struggle to distinguish between these feature types~\citep{zietlow2022leveling}, and tend to indiscriminately modify parameters that contribute to bias. While this approach may successfully eliminate bias features, it often inadvertently impacts core features, thereby improving fairness at the cost of degrading discriminative performance.  
\textcolor{revisecolor}{To tackle this challenge, we propose a targeted debiasing method that operates at the parameter level. Specifically, we differentiate model parameter updates, during debiasing, with respect to their importance to both bias and core features. This allows for a precise intervention that mitigates bias by adjusting the most influential parameters while minimizing perturbations to the parameters that encode core features.}

\textcolor{mycolor}{
We propose an efficient and effective debiasing framework, called Soft-Mask Weight Fine-Tuning (SWiFT). Unlike pre- or in-processing debiasing methods, SWiFT requires only a few epochs of fine-tuning instead of full model retraining. Furthermore, our approach uses only a small external dataset \textcolor{revisecolor}{(in our experimental work, no more than $25\%$ of the original training set size)}, thus eliminating the need for extensive or often inaccessible training data -- a common practical limitation. Compared to post-processing methods, SWiFT effectively preserves core features for prediction while removing bias features, yielding superior debiasing and discriminative performance across diverse metrics. As illustrated in Figure \ref{fig:method},  SWiFT operates under the following procedure: given a pre-trained biased model, the first step is to generate a parameter-wise \textit{soft-mask}, which quantifies each network parameter's relative contribution to bias versus prediction performance. During debiasing, this soft-mask constrains gradient flows during backpropagation, ensuring minimal changes to parameters important for prediction and preserving core features. 
Next, SWiFT employs a two-step debiasing strategy: firstly, it fine-tunes the feature extractor parameters with gradient flows modulated by the soft-mask to remove bias from features; 
\textcolor{revisecolor}{secondly, to eliminate bias from incorrect feature compositions within the classification head, we partially re-initialize its parameters to erase the weights of previously learned, bias features. The entire head is then fine-tuned to learn a robust feature combination with only the core features.}
Our main contributions are summarized as follows:}
\begin{itemize}

    \item [$\bullet$] 
    \textcolor{mycolor}{We propose a novel debiasing framework capable of mitigating bias and improving fairness and generalizability, without compromising model discriminative performance. Our framework circumvents expensive full model retraining and 
    original training data access requirements.}
    
    \item[$\bullet$] \textcolor{mycolor}{
    A soft-mask generation technique that quantitatively measures 
    network parameter contributions towards bias features {vs.}~core features, enabling targeted debiasing through differentiated parameter updates. To the best of our knowledge, 
    our work is the first to employ model parameter soft-masking 
    to the debiasing task. }
    
    \item[$\bullet$] \textcolor{mycolor}{We propose a two-step fine-tuning strategy that is model-agnostic and effectively removes model bias through only a few 
    epochs of fine-tuning on a small external dataset.}
    
    \item[$\bullet$] Experiments on four \textcolor{revisecolor}{out-of-distribution (OOD)} dermatological disease and two OOD chest X-ray datasets 
    demonstrate that SWiFT outperforms nine \textcolor{revisecolor}{state-of-the-art (SOTA)} methods, with respect to both fairness and discriminative capabilities.
    
\end{itemize}
\textcolor{mycolor}{A preliminary version of this work, named BMFT~\citep{xue2024BMFT}, was presented at the MICCAI FAIMI Workshop 2024. In this journal extension we include: i) four methodological modifications that improve generalization, robustness, and effectiveness over our previous method. First, we introduce a novel soft-mask generation module that enables more adaptive and fine-grained model updates. Unlike the previous hard-mask, which optimizes parameters based on a predefined threshold and freezes others, soft-mask does not fully block parameters and keeps most parameters trainable. This provides the network with greater capacity to update parameters towards core features and removes the need for exhaustive threshold tuning. Second, we propose a partial classification head re-initialization strategy. It preserves well-learned core features that facilitates feature re-combination with faster convergence speed. Third, we improve the loss function with dynamic balancing between classification loss and fairness constraints, enhancing adaptability across different tasks and datasets. Finally, we extend SWiFT to handle multi-attribute bias removal scenarios. Modifications are accompanied with new experiments which demonstrate the respective improvements compared to BMFT and other baselines. ii) New experimental validation. 
We 
perform new experiments with 5-fold cross-validation, introduce new experimental domains 
with chest-xray datasets, and evaluate SWiFT on two additional network architectures, EfficientNet-B3 and DenseNet-121, to validate its robustness and generalization to different scenarios. Lastly, we conduct comprehensive ablation studies to analyze the contribution of each module, the role of the external dataset, and the sensitivity to hyperparameters. iii) Additional experiments analysis: we provide an in-depth analysis of the experimental results with additional quantitative evaluations and qualitative insights through Class Activation Maps (CAMs).} 

\section{Related Work}
In this section, we firstly review fairness challenges in medical imaging classification tasks. We next provide a brief survey on recent debiasing methods proposed to tackle these challenges, and finally discuss existing mask fine-tuning techniques, highlighting how our soft-mask-based debiasing approach builds upon and extends previous work. 
\label{sec:lit}
\subsection{Bias and Fairness in Medical Imaging}
Bias and fairness are widely reported in medical imaging analysis~\citep{luo2024harvard, wang2017chestx, zhang2022improving}. 
Bias in ML systems leads to unfair decision-making, thus undermining model fairness. 
Bias can arise from various sources, including label noise, data imbalance, and spurious correlations~\citep{mehrabi2021survey}. 
A critical concern is bias amplification, where biases in the training data are amplified by model predictions during deployment~\citep{duttfairtune}. 
This not only compromises model fairness and generalization, but also risks exacerbating social discrimination~\citep{mehrabi2021survey}.
Achieving fairness in medical AI is thus both a technical challenge and an ethical imperative. 
Fairness in ML systems is often measured through \emph{group fairness}~\citep{mehrabi2021survey}, which aims to ensure similar average outcomes across demographic groups. Common metrics for group fairness include statistical parity difference (SPD)\citep{marcinkevics2022debiasing}, equal opportunity (EO)~\citep{hardt2016equality}, and equalized odds (EOdds)\citep{hardt2016equality}.
However, optimizing for group fairness can lead to a reduction in individual fairness, often resulting in decreased prediction accuracy~\citep{zietlow2022leveling}.
Therefore, debiasing is inevitably a multi-objective challenge that seeks to simultaneously achieve good discriminative performance and good fairness.

\subsection{Debiasing Methods}
\label{sec:lit:methods}
Existing bias mitigation methods in medical imaging can be categorized into three groups: \textit{pre-processing, in-processing}, and \textit{post-processing}. 
\textit{Pre-processing} methods debias the training data before the training process. \textcolor{revisecolor}{Current approaches include either transforming data representations to eliminate correlations between feature representations and sensitive attributes~\citep{zhang2022improving},~\citep{yuan2023edgemixup} or augmenting data distributions to equalize the training distribution across different demographic groups~\citep{oguguo2023comparative},~\citep{zietlow2022leveling}.} Pre-processing methods often require extensive data processing efforts, and commonly used data augmentation strategies (e.g.~color permutation~\citep{park2022fair}) may not be suitable for certain medical modalities (e.g.~CT scans). 
\textcolor{revisecolor}{\textit{In-processing} methods address fairness during the model training process, typically by modifying the loss function to regularize the model and mitigate bias. For example, \citet{zafar2017fairness} 
incorporate bias-specific regularization terms in the loss function.~\citet{jung2023re} employed a classwise distributionally robust optimization framework to mitigate group disparities.~\citet{zhaozero} used adversarial learning to remove sensitive information. In-processing methods cannot explicitly protect the underrepresented group when enforcing the fairness constraints, often resulting in accuracy drops for both groups. Additionally, both pre- and in-processing methods necessitate access to the original training data and require model retraining, limiting their efficiency in real-world scenarios.}
\textcolor{revisecolor}{\textit{Post-processing} techniques, in contrast, are performed after training and 
achieve fairness by modifying model predictions or parameters.~\citet{oguguo2023comparative} achieve fairness by flipping unfavorable predictions of the minority group on instances where the model exhibits high uncertainty.~\citet{marcinkevics2022debiasing} and~\citet{wu2022fairprune} mitigate unfairness by pruning parameters based on their contribution to bias.} However, debiasing by simply changing specific sample predictions or removing neurons may be problematic. Such approaches can be unfair to certain individuals and may cause the model to lose information about core features, leading to unsatisfactory trade-offs between accuracy and fairness~\citep{chen2024fast}.

To address the outlined challenges, 
our method applies fine-tuning over only a few epochs using a small external dataset, without the need to modify the data or retrain the model. Our approach considers both prediction and bias constraints by estimating model parameter contributions towards core and bias features, resulting in high overall accuracy and fairness.

\subsection{Mask Fine-tuning}
\label{sec:lit:finetune}
Fine-tuning a model that has been pre-trained on a large dataset towards a specific downstream task is common practice in machine learning~\citep{bitfit}. Leveraging the knowledge embedded in pre-trained models often enables downstream tasks to be learned with significantly less data compared to training from scratch. Mask fine-tuning methods are a family of techniques designed to improve the fine-tuning process by controlling parameter updates according to their importance on the objective. Current approaches, such as Parameter Efficient Fine Tuning~\citep{bitfit, duttfairtune}, employ the binary-mask (hard-mask) to select a subset of parameters to update while freezing the rest. Hard-masks typically require extensive hyperparameter tuning to determine the optimal selection rate of parameters for fine-tuning versus freezing~\citep{wu2022fairprune, duttfairtune}. Moreover, we note that hard-masks provide only a relatively blunt tool to differentiate the contributions of specific parameters towards the objective. \citet{konishi2023parameter} proposed the use of soft-masks in continual learning tasks, which are flexible and efficient. We investigate a soft-mask fine-tuning strategy to mitigate bias and improve fairness. Overall, our method demonstrates the superiority of soft-mask over binary-mask in terms of both fairness and accuracy. 

\section{Methodology}
\label{sec:method}
The core idea of our method, see Figure~\ref{fig:method}, is to discriminately control model parameter updates based on their relative individual contributions to both bias and predictive performance.  
Key to our method is the soft-mask generation, which regulates model parameter updates during debiasing. The debiasing process consists of a two-step fine-tuning strategy that uses the soft-mask to debias the model feature extractor and the classification head, sequentially. 
Before we proceed to detail our method we begin with preliminaries.

\subsection{Preliminaries}
\label{sec:method:prelim}
\noindent \textbf{Problem Formulation} We study a supervised image classification problem where we will impose fairness considerations. We define disjoint training, validation, and test datasets such that $\mathcal{D}_\mathrm{train}  \cup \mathcal{D}_\mathrm{valid}  \cup \mathcal{D}_\mathrm{test} = \{\mathbf{x}_{i}, y_{i}, a_{i}\}$, where $\mathbf{x}_{i}$ denotes input image $i$ with class label $y_{i}$, and $a_{i}$ is a 1D binary vector that represents the presence or absence of sensitive attributes for sample $i$ (e.g.~skin tone, gender, age). 
In this work we assume binary prediction targets and sensitive attributes (i.e.~$y_{i}, a_{i} \in \{0, 1\}$). Let $f_{\theta}(\cdot)$ denote a biased (i.e.~lacking in group fairness) model parameterized by $\theta$. We assume a decomposable model consisting of a feature extractor $E(\cdot)$ and a classification head $C(\cdot)$, which is pre-trained on the original training data $\mathcal{D}_\mathrm{train}$. 

Our goal is to debias $f_{\theta}(\cdot)$, as measured by existing fairness metrics such as Statistical Parity Difference (SPD)~\citep{dwork2012fairness} and Equalised Odds (EOdds)~\citep{hardt2016equality}, by fine-tuning for only a few epochs with the external dataset $\mathcal{D}_{e}$. With this procedure, we aim to enhance the model's reliance on core features and further improve its generalizability, which can be evidenced by improved AUC on various OOD test datasets $\mathcal{D}_\mathrm{test}$ that share the same sensitive attribute $a_{i}$.

\noindent \textbf{External Dataset Preparation} The external dataset can be readily constructed from labeled data that shares the same task categories as the training data $\mathcal{D}_\mathrm{train}$. 
As a proof of concept, our experimental work 
constructs $\mathcal{D}_{e}$ from each task's respective validation dataset.
The external dataset must contain samples both with and without biased features. This enables the debiasing method to accurately identify biased representations within the pre-trained model and subsequently mitigate them during model fine-tuning.
 To avoid introducing new sources of biases (i.e., group attribution bias, label bias) from $\mathcal{D}_{e}$, we employ common group-balancing strategies~\citep{zhang2022improving, mao2023last} to prepare our external dataset. Specifically, we keep all data from the smallest attribute group, and subsample the data from the other group to the same size and classification label ratios. \textcolor{revisecolor}{Therefore, our method does not require the same label or sensitive distributions between the final external dataset and the training/test data; in fact, it leverages the balanced distribution across sensitive groups within $\mathcal{D}_{e}$ to effectively debias the model.} 
%


\subsection{Soft-Mask Generation}
\label{sec:method:maskgen}
Model parameters contribute non-uniformly to bias features that affect fairness and core features that drive predictive performance~\citep{wu2022fairprune, duttfairtune}. Therefore, parameters require differentiated adjustment during debiasing. To address this, we define a soft-mask that assigns larger updates to parameters contributing primarily to bias and smaller updates that preserve features crucial for prediction. In this way, the core features are retained 
while bias is mitigated. To construct our mask, we calculate the respective per-parameter importance according to relevant prediction and bias functions. 

\noindent \textbf{Prediction function} We adopt Weighted Binary Cross Entropy (WBCE~\citep{xue2024BMFT}), 
which is a robust cross-entropy term w.r.t.~class imbalance -- a common trait of medical imaging datasets~\citep{bevan2022skin, puyol2021fairness}:
\textcolor{revisecolor}{
\begin{equation}
\resizebox{.9\hsize}{!}{$
   \mathcal{L}_{\mathrm{WBCE}}=\sum_i \left(-\frac{N_n}{N_n+N_p}\mathbf{y}_i\log(p_i)-\frac{N_p}{N_n+N_p}(1-\mathbf{y}_{i})\log(1-p_i)\right).
\label{bce_loss}
$}
\end{equation}}
\noindent \textbf{Bias function} We use the differentiable proxy of EOdds~\citep{hardt2016equality}:
\begin{equation}
    \mathcal{B}(f_{\theta}, \mathcal{D}_{e}) = tpr + fpr,
\label{eq:bias_EOdds}
\end{equation}
where 
\textcolor{revisecolor}{
\begin{equation}
\resizebox{.9\hsize}{!}{$
    tpr = \left|
    \frac{\sum_{i}(1\!-\!a_{i})y_{i}\log(f_{\theta}(x_{i}))}{\sum_{i}(1\!-\!a_{i})y_{i}} \!-\! \frac{\sum_{i}a_{i}y_{i}\log(f_{\theta}(x_{i}))}{\sum_{i}a_{i}y_{i}}\right|\raisebox{-2ex}{,}
$}
\end{equation}
\begin{equation}
\resizebox{.9\hsize}{!}{$
    fpr = \left|
    \frac{\sum_{i}\!(1\!-\!a_{i})(1\!-\!y_{i})\log(f_{\theta}(x_{i}))}{\sum_{i}(1\!-\!a_{i})(1\!-\!y_{i})}\!-\! \frac{\sum_{i}a_{i}(1\!-\!y_{i})\log(f_{\theta}(x_{i}))}{\sum_{i}a_{i}(1\!-\!y_{i})}\right|\raisebox{-2ex}{,}
$}
\end{equation}
}
 where $f_{\theta}(x_{i})$ provides a predictive probability for sample $x_{i}$ under a binary classification task. \textcolor{revisecolor}{EOdds requires that a model's predictions are independent of sensitive group memberships with different sensitive groups having the same false positive rates and true positive rates. }\textcolor{revisecolor}{The EOdds is defined for binary sensitive attributes. To extend its application to multi-attribute scenarios, we employ a targeted optimization strategy. Specifically, we first identify the attribute groups with the highest and lowest Area Under the Curve (AUC) scores to serve as proxies for the most advantaged and disadvantaged groups. We then apply the EOdds bias function to this pair of groups. The effectiveness of this approach is evaluated in Section \ref{sec:discussion:multi-attribute}.}

\noindent \textbf{Estimating parameter importance} Previous work~\citep{foster2024fast, kirkpatrick2017overcoming} has demonstrated that the parameter importance towards an objective function can be effectively calculated using the Fisher Information Matrix (FIM). Given a probability density function (PDF) $p(\mathcal{D}_{e}|\theta)$, the FIM over dataset $\mathcal{D}_{e}$ is defined by: 
\begin{equation}
\small
    \mathcal{I}(\theta, \mathcal{D}_{e}) = - \mathbb{E}_{\mathcal{D}_{e}}\bigtriangledown^{2}_{\theta} \log p(\mathcal{D}_{e}|\theta) 
    \label{eq:OrgFIM}
    .
\end{equation}
The full FIM is an $n\times n$ matrix, typically making the calculation computationally expensive. Thus we follow~\citet{foster2024fast, kirkpatrick2017overcoming} and further approximate the FIM using its diagonal values, as given by:
\begin{equation}
\small
    \mathcal{I}(\theta, \mathcal{D}_{e}) = -\mathbb{E}\!\left[\!\left(\frac{\partial \log p(\mathcal{D}_{e}|\theta)}{\partial\theta}\!\right)\!\left(\!\frac{\partial \log p(\mathcal{D}_{e}|\theta)}{\partial\theta}\!\right)^\top\!\right]\raisebox{-2ex}{.}
    \label{eq:FIM}
\end{equation}

 Given a pre-trained model $f_{\theta}$ and the external dataset $\mathcal{D}_{e}$, the log-likelihoods of the prediction PDF and bias PDF are simply the negative of the WBCE function and bias function in Eq.~\eqref{eq:bias_EOdds}, respectively. Therefore, the parameter importance, in terms of influencing prediction accuracy, is given by:
\begin{equation}
\small
    \mathcal{I}_{l}(\theta, \mathcal{D}_{e}) = \mathbb{E}\!\left[\!\left(\frac{\partial \mathcal{L}_\mathrm{WBCE}(\mathcal{D}_{e}|\theta)}{\partial\theta}\!\right)\!\left(\!\frac{\partial\mathcal{L}_\mathrm{WBCE}(\mathcal{D}_{e}|\theta)}{\partial\theta}\!\right)^\top\!\right]
    \label{eq:lossimp}.
\end{equation}
Whereas, the parameter importance in terms of influencing model bias can be computed by:
\begin{equation}
\small
    \mathcal{I}_{b}(\theta, \mathcal{D}_{e}) = \mathbb{E}\!\left[\!\left(\frac{\partial \mathcal{B}(\mathcal{D}_{e}|\theta)}{\partial\theta}\!\right)\!\left(\!\frac{\partial\mathcal{B}(\mathcal{D}_{e}|\theta)}{\partial\theta}\!\right)^\top\!\right]
    \label{eq:biasimp}.
\end{equation}

\begin{figure}[t]
    \centering
    \includegraphics[width=1\columnwidth]{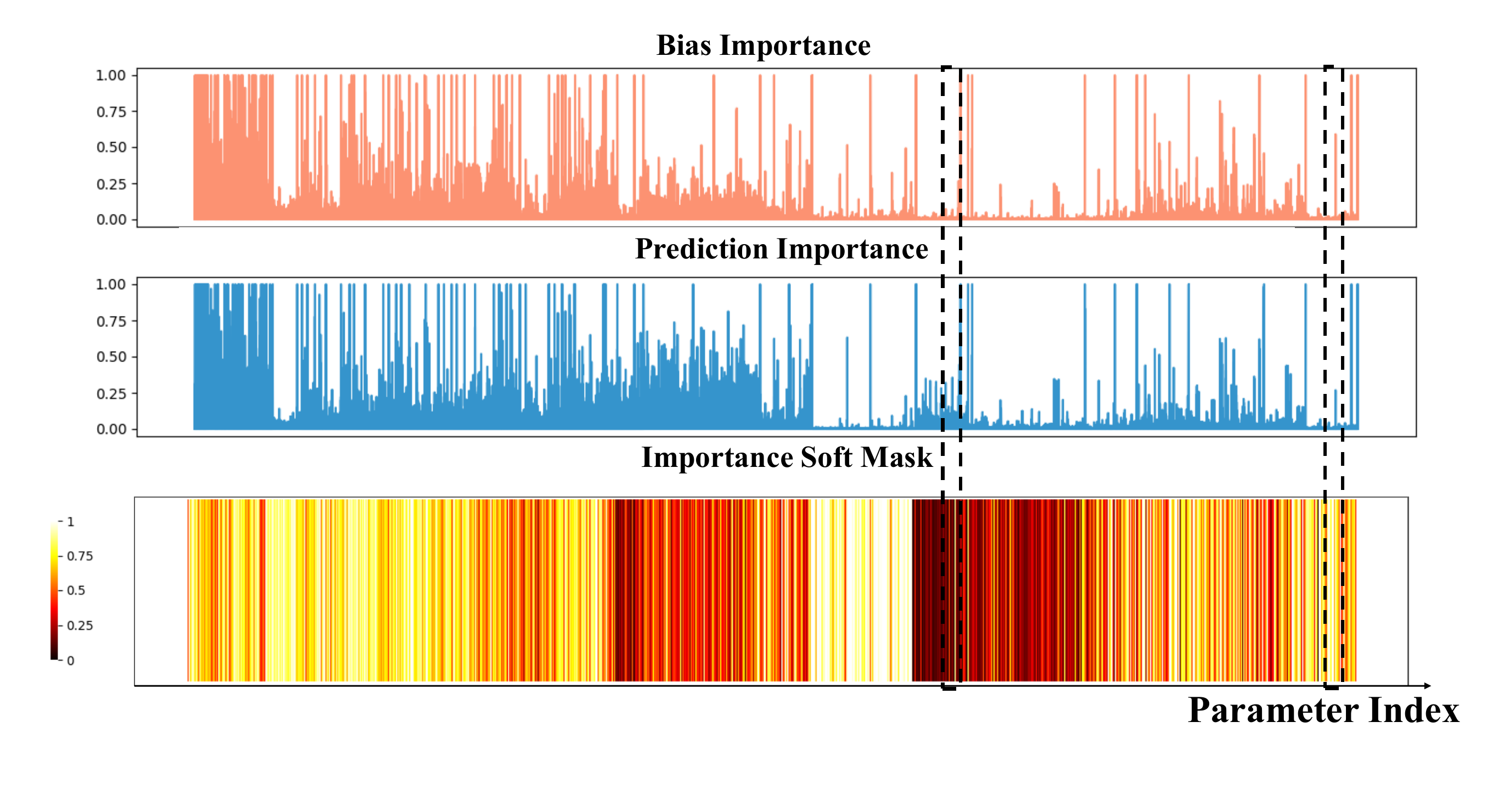}
    \caption{Soft-mask generation process. Each model parameter is assessed for (a) bias importance and (b) prediction importance for dark and light skin tone groups on the Fitzpatrick-17k dataset. The x-axis is the index of parameters of the ResNet-50. The soft-mask (c) is created using Eq.~\eqref{eq:mask}. The black dotted boxes highlight those parameters that show different importance towards prediction accuracy and bias: parameters with high prediction but lower bias importance (left dotted box), and parameters with high bias but lower prediction importance (right dotted box).}
    \label{fig2}
\end{figure}

\noindent \textbf{Mask Construction} Having defined estimates for parameter importance, with respect to both predictive accuracy and bias, we design our weight soft-mask according to the relative ratio between the introduced importance terms.  
We denote each continuous scalar element of the weight mask as $M_i$, where $i$ is the weight index. An element of the mask is defined as:
\begin{equation}
    \mathcal{M}_{i} = \left| \mathrm{\tanh} \left( \frac{\text{Norm}(\mathcal{I}_{i, b})}{\text{Norm}(\mathcal{I}_{i, l})} \right)\right|,
\label{eq:mask}
\end{equation}
where
$\mathcal{I}_{i, l}$ and $\mathcal{I}_{i, b}$ are the $i$-th diagonal elements of the FIMs in Eq.~\eqref{eq:lossimp} and Eq.~\eqref{eq:biasimp}, respectively, 
and $\text{Norm}(\cdot)$ denotes min--max normalization. 
Normalizing gradients with respect to each network layer helps us avoid discrepancies due to large differences in gradient magnitude. \textcolor{revisecolor}{The components of Eq.~\eqref{eq:mask} serve specific purposes. First, since the two importance terms may have different numerical scales, we normalize them with respect to each network layer. This ensures their ratio provides a meaningful measure of a parameter's relative contribution to bias versus prediction. Our choice is empirically validated in Section~\ref{sec:discussion:ablate:normal}, where we show that min-max normalization yields superior results compared to z-score normalization for this task. Second, the raw ratio of these terms is unbounded. We therefore apply the hyperbolic tangent (tanh) function to introduce non-linearity and map the unbounded input to a finite interval.  Finally, since the magnitude of the gradient rather than its sign indicates a strong importance (i.e., values near $-1$ and $+1$ are both significant), we take the absolute value. This ensures that each mask element $M_{i}$ lies within the range $[0,1]$.}
Due to this construction, a \emph{higher value of $M_i$ represents a weight that is predicted to have larger importance for attribute bias and a less significant impact on predictive accuracy. } The parameters with higher mask values will constitute the significant updates during fine-tuning. 
In the following section, we provide a strategy that uses this signal to remove the detrimental influence of model bias, while retaining information embedded in core features. 
An illustrative example of bias importance, prediction importance, and the resulting soft-mask is provided in Figure~\ref{fig2}.

\subsection{Two-Step Fine-Tuning}
\label{sec:method:twostep}
Model bias typically originates from two key sources~\citep{le2023last}: (1) the \textit{entanglement} between harmful bias features (typically spurious or irrelevant) and useful core (causal) features, within the feature extractor; and (2) \textcolor{revisecolor}{the \textit{incorrect composition} of representations in the classification head (i.e., bias features are highly weighted in the classification head, causing the final predictions to rely on bias).} 
Our method establishes a two-step fine-tuning process to address the 
two key sources sequentially. 

\noindent \textbf{Feature Extractor: Soft-Mask Fine-Tuning}
\label{sec:method:twostep:extractor}
Core features that exist in pre-trained models are often entangled with bias features~\citep{le2023last}. To address this, our first fine-tuning step debiases the feature extractor $E(\cdot)$ while keeping the classification head $C(\cdot)$ frozen. The goal of this fine-tuning process is two-fold: to remove bias features that may compromise fairness and to preserve core features essential for accurate prediction. We achieve this goal by optimizing the feature extractor with a bias-specific loss function, while constraining updates of prediction-important parameters. 
Specifically, we fine-tune the feature extractor on $\mathcal{D}_{e}$ using a loss function that combines the WBCE loss $\mathcal{L}_{\mathrm{WBCE}}$ with a fairness constraint (Eq.~\ref{eq:bias_EOdds}):
\begin{equation}
\small
\mathcal{L}=\beta\mathcal{L}_{\mathrm{WBCE}}(f_{\theta}, \mathcal{D}_{e}) + (1-\beta) \mathcal{B}(f_{\theta}, \mathcal{D}_{e}),
    \label{eq:repairobj}
\end{equation}
where $\mathcal{B}(f_{\theta}, \mathcal{D}_{e})$ 
represents our objective relating to 
reducing bias. \textcolor{revisecolor}{The hyperparameter $\beta \in [0, 1]$ explicitly balances the prediction accuracy and fairness objectives. $\beta$ is fixed within each step of our two-step fine-tuning pipeline, but may differ between steps. In this first step, we set $\beta$ to a small value $\epsilon$ to prioritize feature debiasing during feature extractor fine-tuning. }During the backward pass, we modify the gradients of all parameters in the feature extractor using the soft-mask. This mask modulates each parameter's update according to its relative bias-prediction importance ratio:
\begin{equation}
    g_{i}^{\prime} = \mathcal{M}_{i}g_{i},
    \label{eq:maskupdate}
\end{equation}
where $g_{i}$ and $g_{i}^{\prime}$
represent the original and modified gradients of the $i^{\text{th}}$ model parameter $\theta_{i}$, respectively.
This soft-masked gradient adjustment ensures that parameters critical for prediction accuracy are updated less aggressively, while bias-related parameters are targeted for more significant modification.

\noindent \textbf{Classification Head: Re-initialization and Fine-tuning}
\label{sec:method:twostep:reinit}
The previous fine-tuning step reduces bias originating from core and bias feature entanglement in the feature extractor. We introduce here our second step to re-combine debiased features within the classification head. 
Since the feature extractor already contains well-learned representations, fine-tuning the classification head is known to converge very quickly,
often before model parameters can be sufficiently updated towards new or modified objectives~\citep{li2020rifle}.
In our setting, this results in
classification decisions remaining heavily influenced by original features rather than debiased features~\citep{zhang2017defense}, limiting fairness and accuracy gains.
To address this issue, we propose parameter re-initialization as an effective strategy to push significant and meaningful updates in the classification head, thus promoting the learning of correct combinations of debiased core features~\citep{ramkumareffectiveness,kirichenkolast}. 

Recent works~\citep{xue2024BMFT,mao2023last} suggest strategies that re-initialize the entire classification head $C(\cdot)$. This risks losing all discriminative ability of the pre-trained model where recovery would require retraining, typically not achievable by small epoch counts. Therefore, we alternatively re-initialize only a portion of parameters, based on the mask calculated using Eq.~\ref{eq:mask}.
Parameters with high mask values, indicating their disproportionately strong influence on the bias features, are re-initialized to zero. This zero re-initialization strategy is inspired by \citet{le2023last} where it was shown that, for an unbiased model trained on a balanced dataset, most classification head parameters converge to zero. In our setting, re-initializing bias-important parameters to zero facilitates rapid convergence towards an unbiased state where it is understood that subsets of parameters can be initialized to zero without causing training degeneracy~\citep{zhaozero}. 
The re-initialization threshold $\gamma$ is determined by the mean value of the mask calculated on the classification head parameters. The partial re-initialization process is shown as Eq.~\eqref{eq:partial_reinit}.


\begin{equation}
\small
\theta_{i}^\prime = \left\{
\begin{array}{ll}
0,             & \text{if} \ \ \mathcal{M}_{\theta_{i}} \geq \gamma \\
\theta_{i},    & \text{otherwise}, \\
\end{array} \right.
\label{eq:partial_reinit}
\end{equation}
Finally, we fine-tune the re-initialized classification head while keeping the feature extractor frozen, using the 
Eq.~\eqref{eq:repairobj} objective. \textcolor{revisecolor}{Different from the first step, we typically set a much larger hyperparameter value of \mbox{$\beta$ = 
$1-\epsilon$} at this stage, to place more emphasis on retaining the model's discriminative capability.}

\section{Experimental Setup}
\label{sec:experiments}
We next detail our experimental work 
where the proposed method is validated under two inference tasks: skin lesion classification and chest X-ray classification. In the following subsections we introduce the considered tasks, provide implementation details and evaluation settings. 

\subsection{Datasets and Pre-processing}
\label{sec:experiments:data}
In both tasks, we perform a five-fold cross-validation to evidence the efficacy and statistical significance of our method. The original dataset is split into training set and validation set with an $80{:}20$ ratio, ensuring no overlap. The model is pre-trained on the training dataset and was debiased via fine-tuning on the external dataset (i.e.~constructed from the validation dataset). \textcolor{revisecolor}{The debiased model is then tested on various out-of-distribution (OOD) test datasets, which may exhibit distributional shifts in both disease labels and sensitive attributes, relative to the training data.}

\subsubsection{Skin Lesion Classification}
\label{sec:experiments:data:skin}
We use the International Skin Imaging Collaboration (ISIC) Challenge Dataset for pre-training and fine-tuning.
We treat ``melanoma" as the negative class and all other lesions as the positive class.
We combine 2017~\citep{codella2018skin}, 2018~\citep{codella2019skin}, 2019~\citep{tschandl2018ham10000,hernandez2024bcn20000} and 2020~\citep{rotemberg2021patient} ISIC challenge data, and remove duplicate images across different years, totaling 56,863 images. Among these, 45,489 images are used for training, while 11,374 images are reserved as the validation dataset in each fold. 
We consider two sensitive attributes: skin tone and gender.
We maintain the same skin tone annotation as~\citet{bevan2022skin} for training data. 
To build the group-balanced external dataset $\mathcal{D}_{e}$, we select c.~3,600 images for skin tone and c.~11,000 for gender. 
All images are pre-processed with center-cropping and resizing to size $256{\times}256$.

For testing, our study employs four OOD datasets with skin images for melanoma detection.
Fitzpatrick-17k~\citep{groh2021evaluating} contains 16,577 images across six skin tone levels, which we group into light (1-3) and dark (4-6) categories.  
DDI~\citep{daneshjou2022disparities} offers 656 images with skin tone annotations. 
Interactive Atlas of Dermoscopy (Atlas)~\citep{lio2004interactive} and PAD-UFES-20~\citep{pacheco2020pad} contain gender labels with 1,011 images and 2,298 images, respectively. 
\subsubsection{Chest X-ray Classification}
\label{sec:experiments:data:xray}
We train a model on the large-scale chest X-ray dataset MIMIC-CXR~\citep{johnson2019mimic}. We use only frontal view images, and resize to $224{\times}224$ pixels. We consider a binary classification problem of ``pneumothorax'' and ``No Findings'', where these classes are treated as positive and negative, respectively. The problem definition is consistent with previous chest X-ray fairness research~\citep{seyyed2020chexclusion,zhang2022improving}. In each cross-validation fold, 61,502 images are randomly selected for training, while the remaining 15,376 images are used as the validation set. 
We focus on two sensitive attributes (gender and age) since the groups of these attributes are previously shown to have disparate classification outcomes~\citep{wang2017chestx,zhang2022improving}. For age, we categorize individuals into a younger group ($\leq65$ years old) and an older group ($>65$ years old), following established definitions in~\citep{singh2014defining}. 14,950 images for gender and 10,360 images for age are selected to build the external dataset $\mathcal{D}_{e}$ for fine-tuning.

We evaluate our method on two OOD datasets; Chexpert~\citep{irvin2019chexpert} and Chest-Xray8 (NIH)~\citep{wang2017chestx}.
We select CXRs with ``Pneumothorax'', ``No Findings'' labels from Chexpert and NIH, which are 21,040 and 10,815 images, respectively.
Both datasets include sensitive attributes: gender (Female and Male) and age (0--80), making them suitable for assessment of fairness {and accuracy metrics.} 

\subsection{Implementation}
\label{sec:experiments:imp}
We conduct experiments in PyTorch using one NVIDIA A100 40GB GPU. For skin lesion classification, We apply ImageNet-pretrained ResNet50 and Efficient B3 as the model backbone. The models are trained for 200 epochs using an SGD optimizer with a batch size 128 and a learning rate 1e-4. For chest X-ray classification, we choose ImageNet-pretrained ResNet50 and DenseNet-121 as the backbone. The models are trained for 100 epochs using an Adam optimizer with a batch size of 128 and a learning rate of 1e-4. The pre-training process for both tasks follows previous fairness research~\citep{bevan2022skin, mao2023last, zhang2022improving, marcinkevics2022debiasing}. 
For debiasing the model, we fine-tune the skin pre-trained model and the chest X-ray pre-trained model for an additional 20, 10 epochs, 
respectively, which are 10\% of the initial training duration. 
Equal epoch counts are used for 
both stages of two-step fine-tuning. 
The fine-tuning process shares the same optimizer and batch size as the training process. 
%
%
Data augmentation was applied in both training and fine-tuning, which includes random flipping, random transpose, and $z$-score normalization. 

\subsection{Comparison Method and Evaluation Metrics}
\label{sec:experiments:metrics}
We compare our methods with nine recent SOTA models. The model pre-trained on the training dataset $\mathcal{D}_{\mathrm{train}}$ is our basic \textbf{Baseline}. \textbf{FullFT-RW}~\citep{zhang2022improving} fine-tunes the pre-trained model on the group-balanced external data $\mathcal{D}_{e}$ using only the prediction loss. \textbf{FullFT-Reg}~\citep{cherepanova2021technical} fine-tunes the pre-trained model on the non group-balanced dataset $\mathcal{D}_{val}$ using a combination of prediction loss and fairness constraints in the form of Eq.~\eqref{eq:repairobj}. \textbf{FullFT-FDR}~\citep{le2023last} combines data balancing and fairness constraints, fine-tuning all parameters on the dataset $\mathcal{D}_{e}$ with the loss function specified in Eq.~\eqref{eq:repairobj}. 
 \textbf{LLFT}~\citep{mao2023last} fine-tunes only the last layer of a deep classification model to promote fairness.
Similarly, \textbf{DiffGda}~\citep{marcinkevics2022debiasing}  fine-tunes on an external dataset, using a bias-aware loss function to steer network optimization. \textbf{FairPrune}~\citep{wu2022fairprune} improves fairness by pruning parameters based on weight saliency. \textbf{DiffPrune}~\citep{marcinkevics2022debiasing} prunes parameters based on their contributions to bias. \textbf{BMFT}~\citep{xue2024BMFT} is a hard-mask based fine-tuning method for debiasing (our work, prior to SWiFT). 

To evaluate the discriminative performance of different models, we use the area under the curve (AUC) as a primary performance metric, and statistical parity difference (SPD) (\citep{daneshjou2022disparities}) and equalized odds (EOdds) (\citep{hardt2016equality}) as fairness metrics, similar to previous work~\citep{duttfairtune,chen2024fast,wu2022fairprune}.

\begin{figure*}[ht!]
    \centering
    \begin{subfigure}[b]{0.45\textwidth}
        \centering
        \includegraphics[width=0.9\textwidth]{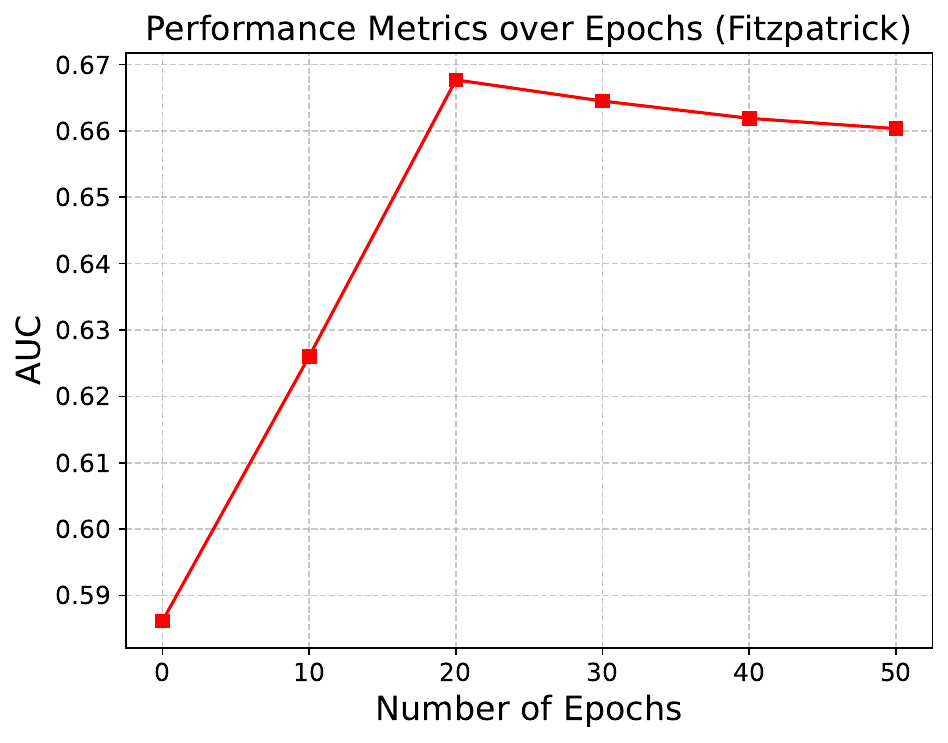}
        \caption{}
        \label{fig:NumOfEp_AUC_Fitz}
    \end{subfigure}
    \hfill
    \begin{subfigure}[b]{0.45\textwidth}
        \centering
        \includegraphics[width=0.9\textwidth]{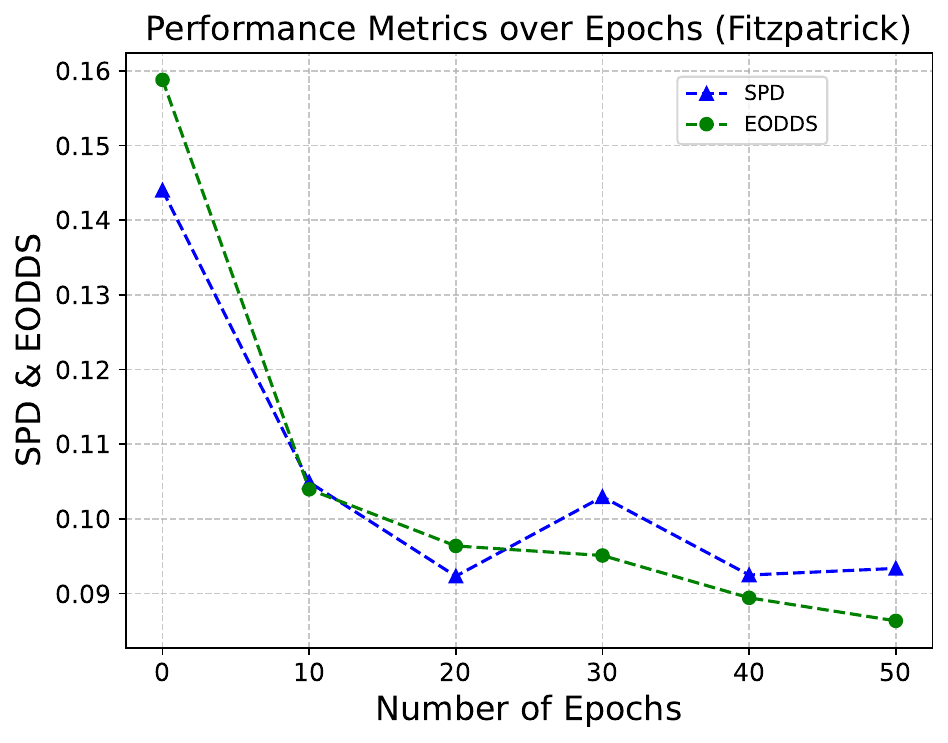}
        \caption{}
        \label{fig:NumOfEp_fair_Fitz}
    \end{subfigure}
    \begin{subfigure}[b]{0.45\textwidth}
        \centering
        \includegraphics[width=0.9\textwidth]{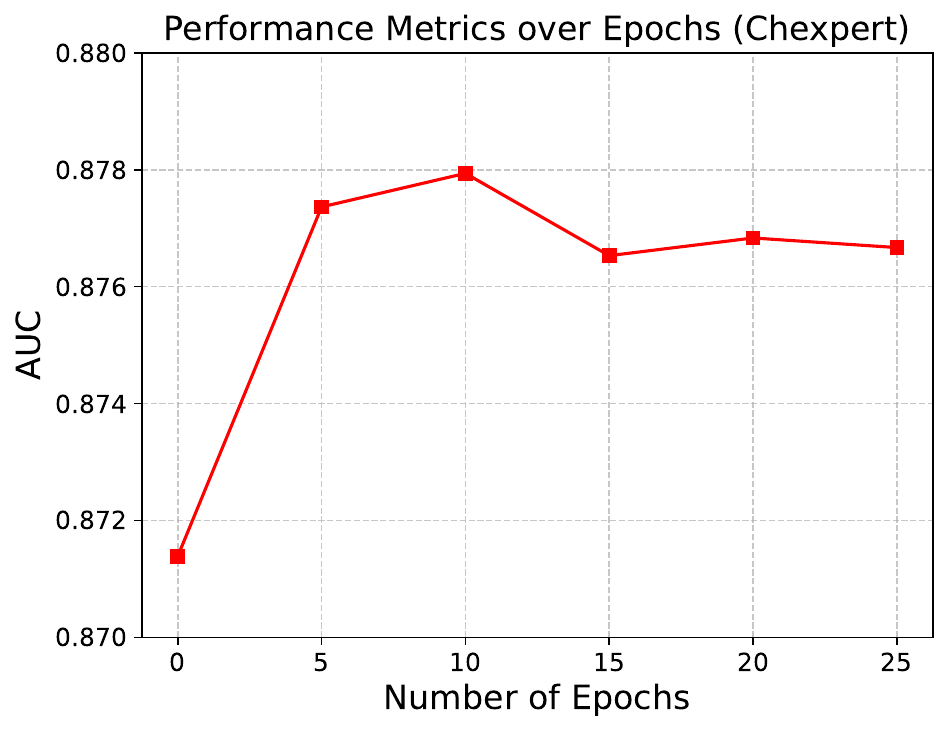}
        \caption{}
        \label{fig:NumOfEp_AUC_chex}
    \end{subfigure}
    \hfill
    \begin{subfigure}[b]{0.45\textwidth}
        \centering
        \includegraphics[width=0.9\textwidth]{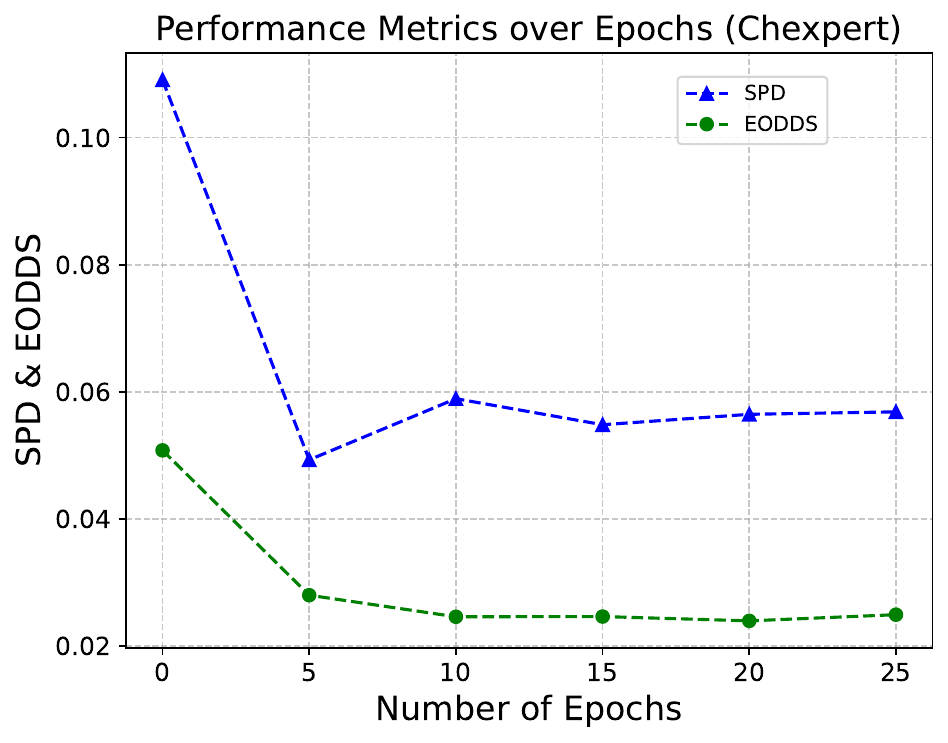}
        \caption{}
        \label{fig:NumOfEp_fair_chex}
    \end{subfigure}

    \caption{Comparison of performance as a function of fine-tuning epochs on skin tone for the skin lesion classification task and age for the chest X-ray classification task. ResNet-50 backbone. (a) and (b) are on the Fitzpatrick-17k dataset for skin tone debiasing, (c) and (d) are on the Chexpert dataset for age debiasing. The performance reaches steady after $20$ epochs for skin tone debiasing and $10$ epochs for age debiasing.}
    \label{fig:epochs_comparison}
\end{figure*}

\subsection{Hyperparameter Validation}
\textcolor{mycolor}{Determining the optimal number of fine-tuning epochs is important for effectively balancing computational efficiency and debiasing performance for our experimental work. We empirically explore the impact of the number of fine-tuning epochs on skin tone debiasing in skin lesion classification and age debiasing in the chest X-ray classification. The results are shown in Figure~\ref{fig:epochs_comparison}. The $0^{th}$ epoch is the pre-trained baseline model. As we can see, both prediction and fairness metrics improve even with only $5\%$ of the original training epochs (i.e., 10 epochs for skin tone debiasing, and 5 epochs for age debiasing), and become steady after $10\%$ of the original training epochs. The debiasing time increases linearly with the number of fine-tuning epochs. Moreover, the AUC tends to decrease as the number of epochs increases, likely due to overfitting during fine-tuning on the small dataset. Based on these findings, we select $10\%$ of the training epochs (i.e., $20$ epochs for skin tone debiasing and $10$ epochs for age debiasing) for subsequent evaluation and comparison with other methods.}

\section{Results}
We report 
performance 
across our investigated experimental scenarios where we aim to answer our primary question: \emph{can model fairness be improved, and discriminative performance maintained, under data and compute frugal regimes?} 
Further, we conjecture that debiased models should generalize effectively across different data distributions and sensitive attributes. Towards exploring this point we evaluate the fairness and predictive accuracy of our approach on several OOD test datasets, each with distinct sensitive attributes.
\label{sec:results}
\subsection{Comparison with SOTA Debiasing Methods}
\subsubsection{Quantitative Results}
\begin{table*}[ht!]
\caption{Comparison of debiasing methods on skin tone and gender. ResNet-50 model pre-trained on ISIC dataset (MEAN$\pm$STD). Results are consistently denoted \textbf{best} and \underline{second best} for all Tables.}\label{tab1}
\centering
{\begin{adjustbox}{width=1\textwidth}
\begin{tabular}{clllllll}
\toprule
 \multirow{2}{*}{Attr.} &  \multirow{2}{*}{Methods} & \multicolumn{3}{c}{Fitzpatrick-17k} & \multicolumn{3}{c}{DDI} \\ \cmidrule(lr){3-5} \cmidrule{6-8}
 &  & \makecell{AUC $\uparrow$} & \makecell{SPD $\downarrow$} & \makecell{EOdds $\downarrow$} & \makecell{AUC $\uparrow$} & \makecell{SPD $\downarrow$} & \makecell{EOdds $\downarrow$} \\ 
 \hline
\multirow{10}{*}{Skin Tone} 
& Baseline &0.586$\pm$0.008& 0.144$\pm$0.008 & 0.159$\pm$0.007 & 0.601$\pm$0.005 & 0.101$\pm$0.020 & 0.113$\pm$0.025\\
& FullFT-RW~\citep{zhang2022improving} & 0.586$\pm$0.009 & 0.146$\pm$0.013 & 0.160$\pm$0.011 & 0.592$\pm$0.008 & 0.126$\pm$0.020 & 0.141$\pm$0.022 \\
& FullFT-Reg~\citep{cherepanova2021technical} & 0.589$\pm$0.006 & 0.118$\pm$0.015 & 0.131$\pm$0.012 & 0.587$\pm$0.006 & 0.116$\pm$0.042 & 0.126$\pm$0.034 \\
& FullFT-FDR~\citep{le2023last} & 0.587$\pm$0.013 & 0.110$\pm$0.015 & 0.123$\pm$0.012 & 0.580$\pm$0.008 & 0.128$\pm$0.040 & 0.128$\pm$0.038 \\
& LLFT~\citep{mao2023last} & 0.571$\pm$0.016 & \underline{0.097}$\pm$0.016 & \underline{0.113}$\pm$0.014 & 0.590$\pm$0.013 & \underline{0.069}$\pm$0.037 & \underline{0.088}$\pm$0.043 \\
& DiffGda~\citep{marcinkevics2022debiasing} & 0.586$\pm$0.008 & 0.144$\pm$0.006 & 0.156$\pm$0.008 & 0.600$\pm$0.005 & 0.106$\pm$0.015 & 0.116$\pm$0.017 \\
& FairPrune~\citep{wu2022fairprune} & 0.560$\pm$0.007 & 0.101$\pm$0.012 & 0.120$\pm$0.015 & 0.564$\pm$0.028 & 0.102$\pm$0.046 & 0.119$\pm$0.055\\
& DiffPrune~\citep{marcinkevics2022debiasing} & 0.589$\pm$0.008 & 0.116$\pm$0.006 & 0.127$\pm$0.006 & 0.598$\pm$0.004 & 0.086$\pm$0.010 & 0.097$\pm$0.009\\
&
BMFT~\citep{xue2024BMFT} & \underline{0.651}$\pm$0.012 & 0.119$\pm$0.013 & 0.114$\pm$0.012 & \underline{0.610}$\pm$0.015 & 0.099$\pm$0.015 & 0.094$\pm$0.020\\
& SWiFT (Ours) & \textbf{0.668}$\pm$0.013 & \textbf{0.092}$\pm$0.028 & \textbf{0.096}$\pm$0.026 & \textbf{0.625}$\pm$0.010 & \textbf{0.063}$\pm$0.030 & \textbf{0.064}$\pm$0.036 \\
\midrule
\multirow{2}{*}{} & \multirow{2}{*}{} & \multicolumn{3}{c}{Atlas} & \multicolumn{3}{c}{PAD} \\ \cmidrule(lr){3-5} \cmidrule{6-8}
 &  & \makecell{AUC $\uparrow$} & \makecell{SPD $\downarrow$} & \makecell{EOdds $\downarrow$} & \makecell{AUC $\uparrow$} & \makecell{SPD $\downarrow$} & \makecell{EOdds $\downarrow$} \\ 
 \hline
 \multirow{9}{*}{Gender}& Baseline & \underline{0.788}$\pm$0.005 & 0.026$\pm$0.008 & 0.031$\pm$0.010 & 0.717$\pm$0.009 & 0.009$\pm$0.005 & 0.036$\pm$0.014 \\
& FullFT-RW~\citep{zhang2022improving} & 0.788$\pm$0.006 & 0.021$\pm$0.007 & 0.022$\pm$0.006 & 0.718$\pm$0.008 & 0.011$\pm$0.008 & 0.035$\pm$0.020 \\
& FullFT-Reg~\citep{cherepanova2021technical}  & 0.780$\pm$0.005 & 0.024$\pm$0.011 & 0.041$\pm$0.018 & 0.703$\pm$0.017 & 0.018$\pm$0.005 & 0.036$\pm$0.030 \\
& FullFT-FDR~\citep{le2023last} & 0.779$\pm$0.006 & 0.030$\pm$0.011 & 0.040$\pm$0.019 & 0.702$\pm$0.018 & 0.012$\pm$0.007 & 0.051$\pm$0.031 \\
& LLFT~\citep{mao2023last} & 0.774$\pm$0.004 & 0.019$\pm$0.006 & 0.024$\pm$0.003 & 0.701$\pm$0.008 & 0.012$\pm$0.006 & 0.031$\pm$0.018 \\
& DiffGda~\citep{marcinkevics2022debiasing} & 0.786$\pm$0.005 & 0.030$\pm$0.013 & 0.035$\pm$0.018 & 0.723$\pm$0.008 & 0.011$\pm$0.008 & \textbf{0.022}$\pm$0.014 \\
& FairPrune~\citep{wu2022fairprune} & 0.691$\pm$0.051 & 0.016$\pm$0.010 & 0.018$\pm$0.010 & 0.619$\pm$0.057 & 0.018$\pm$0.014 & 0.055$\pm$0.019 \\
& DiffPrune~\citep{marcinkevics2022debiasing} & 0.768$\pm$0.008 & 0.006$\pm$0.004 & \underline{0.012}$\pm$0.008 & 0.719$\pm$0.010 & 0.010$\pm$0.013& 0.033$\pm$0.015 \\
& BMFT~\citep{xue2024BMFT} & 0.787$\pm$0.005 & \textbf{0.005}$\pm$0.005 & 0.014$\pm$0.006 & \underline{0.727}$\pm$0.011 & \textbf{0.005}$\pm$0.005& 0.032$\pm$0.032 \\
&
SWiFT (Ours) & \textbf{0.789}$\pm$0.005 & \underline{0.012}$\pm$0.007 & \textbf{0.010}$\pm$0.006 & \textbf{0.728}$\pm$0.009 & \underline{0.006}$\pm$0.005 & \underline{0.031}$\pm$0.004 \\
 \bottomrule
\end{tabular}
\end{adjustbox}
}
\end{table*}

\begin{table*}[ht!]
\caption{Comparison of debiasing methods on skin tone and gender. Efficient-B3 model pre-trained on ISIC dataset.}
\label{tab2}
\centering
{\begin{adjustbox}{width=1.0\textwidth}
\begin{tabular}{clllllll}
\toprule
 \multirow{2}{*}{Attr.} &  \multirow{2}{*}{Methods} & \multicolumn{3}{c}{Fitzpatrick-17k} & \multicolumn{3}{c}{DDI} \\ \cmidrule(lr){3-5} \cmidrule{6-8}
 &  & \makecell{AUC $\uparrow$} & \makecell{SPD $\downarrow$} & \makecell{EOdds $\downarrow$} & \makecell{AUC $\uparrow$} & \makecell{SPD $\downarrow$} & \makecell{EOdds $\downarrow$} \\ 
 \hline
\multirow{10}{*}{Skin Tone} 
& Baseline & 0.625$\pm$0.008 & 0.039$\pm$0.005 & 0.047$\pm$0.007 & 0.600$\pm$0.008 & 0.082$\pm$0.008 & 0.101$\pm$0.014\\
& FullFT-RW~\citep{zhang2022improving} & 0.625$\pm$0.009 & 0.049$\pm$0.007 & 0.053$\pm$0.012 & 0.600$\pm$0.009 & 0.112$\pm$0.019 & 0.121$\pm$0.022 \\
& FullFT-Reg~\citep{cherepanova2021technical} & 0.610$\pm$0.006 & 0.029$\pm$0.009 & 0.032$\pm$0.012 & 0.594$\pm$0.005 & 0.089$\pm$0.004 & 0.119$\pm$0.007 \\
& FullFT-FDR~\citep{le2023last} & 0.616$\pm$0.008 & 0.030$\pm$0.013&0.029$\pm$0.015	&\underline{0.603}$\pm$0.005 & 0.080$\pm$0.015 & 0.109$\pm$0.022 \\
& LLFT~\citep{mao2023last} & 0.612$\pm$0.008	& \underline{0.014}$\pm$0.009 & 0.028$\pm$0.009	& 0.596$\pm$0.012 & \underline{0.029}$\pm$0.018	& \underline{0.055}$\pm$0.023
 \\
& DiffGda~\citep{marcinkevics2022debiasing} & 0.610$\pm$0.068	&0.019$\pm$0.008& \underline{0.025}$\pm$0.006 &	0.589$\pm$0.018	& 0.041$\pm$0.041 &	0.055$\pm$0.050
 \\
& FairPrune~\citep{wu2022fairprune} &0.571$\pm$0.051	& 0.039$\pm$0.025 & 0.047$\pm$0.018	& 0.594$\pm$0.033 &	0.064$\pm$0.043 & 0.064$\pm$0.035\\
& DiffPrune~\citep{marcinkevics2022debiasing} & 0.591$\pm$0.017	& 0.032$\pm$0.021 & 0.029$\pm$0.016	& 0.587$\pm$0.016 &
0.067$\pm$0.022 & 0.078$\pm$0.020 \\
& BMFT~\citep{xue2024BMFT} & \underline{0.640}$\pm$0.005	& 0.029$\pm$0.013 & 0.035$\pm$0.014	& 0.603$\pm$0.009 &
0.050$\pm$0.005 & 0.070$\pm$0.008 \\
& SWiFT (Ours) & \textbf{0.644}$\pm$0.007 & \textbf{0.013}$\pm$0.001 & \textbf{0.019}$\pm$0.014 & \textbf{0.607}$\pm$0.007 & \textbf{0.019}$\pm$0.013 & \textbf{0.025}$\pm$0.014 \\
\midrule
\multirow{2}{*}{} & \multirow{2}{*}{} & \multicolumn{3}{c}{Atlas} & \multicolumn{3}{c}{PAD} \\ \cmidrule(lr){3-5} \cmidrule{6-8}
 &  & \makecell{AUC $\uparrow$} & \makecell{SPD $\downarrow$} & \makecell{EOdds $\downarrow$} & \makecell{AUC $\uparrow$} & \makecell{SPD $\downarrow$} & \makecell{EOdds $\downarrow$} \\ 
 \hline
 \multirow{10}{*}{Gender}& Baseline & 0.785$\pm$0.003 &	0.028$\pm$0.005 & 0.013$\pm$0.003 & 0.599$\pm$0.017 & \underline{0.009}$\pm$0.006 &	0.029$\pm$0.026 \\
& FullFT-RW~\citep{zhang2022improving} & 0.784$\pm$0.002 & 0.026$\pm$0.012 & 0.010$\pm$0.006 &	0.601$\pm$0.012 & 0.015$\pm$0.010 & 0.023$\pm$0.015 \\
& FullFT-Reg~\citep{cherepanova2021technical}  & \underline{0.787}$\pm$0.004	& 0.035$\pm$0.007 & 0.018$\pm$0.008 &	0.592$\pm$0.012	& 0.012$\pm$0.008 & 0.037$\pm$0.020\\
& FullFT-FDR~\citep{le2023last} & \textbf{0.787}$\pm$0.003 & 0.036$\pm$0.013 & 0.021$\pm$0.008 &	0.594$\pm$0.013	& 0.017$\pm$0.009 & 0.049$\pm$0.032 \\
& LLFT~\citep{mao2023last} & 0.785$\pm$0.005 & 0.035$\pm$0.015 & 0.025$\pm$0.013 &
0.604$\pm$0.039	& 0.026$\pm$0.017 & 0.041$\pm$0.027\\
& DiffGda~\citep{marcinkevics2022debiasing} & 0.786$\pm$0.002 & 0.025$\pm$0.007 & 0.011$\pm$0.006 &	0.594$\pm$0.009	& \textbf{0.007}$\pm$0.006 & 0.033$\pm$0.021\\
& FairPrune~\citep{wu2022fairprune} & 0.747$\pm$0.017 & 0.031$\pm$0.030	& 0.025$\pm$0.013 &	0.551$\pm$0.023	& 0.020$\pm$0.012 & 0.031$\pm$0.022\\
& DiffPrune~\citep{marcinkevics2022debiasing} & 0.741$\pm$0.010	& 0.025$\pm$0.020 & \underline{0.009}$\pm$0.003 &	0.599$\pm$0.017 & 0.010$\pm$0.006 & 0.032$\pm$0.029\\
& BMFT~\citep{xue2024BMFT} & 0.781$\pm$0.012	& \underline{0.009}$\pm$0.006 & 0.024$\pm$0.015 &	\underline{0.624}$\pm$0.015 & 0.011$\pm$0.007 & \underline{0.020}$\pm$0.018\\
& SWiFT (Ours) & 0.785$\pm$0.003 & \textbf{0.009}$\pm$0.005 & 	\textbf{0.006}$\pm$0.004
& \textbf{0.631}$\pm$0.010 & 0.015$\pm$0.006 & \textbf{0.014}$\pm$0.002 \\
 \bottomrule
\end{tabular}
\end{adjustbox}
}
\end{table*}

\begin{table*}[ht]
\caption{\mbox{Comparison of debiasing methods on gender and age attributes. ResNet-50 model pre-trained on MIMIC
datasets.}}\label{tab3}
\centering
{\begin{adjustbox}{width=1.0\textwidth}
\begin{tabular}{clllllll}
\toprule
 \multirow{2}{*}{Attr.} &  \multirow{2}{*}{Methods} & \multicolumn{3}{c}{Chexpert} & \multicolumn{3}{c}{NIH} \\ \cmidrule(lr){3-5} \cmidrule{6-8}
 &  & \makecell{AUC $\uparrow$} & \makecell{SPD $\downarrow$} & \makecell{EOdds $\downarrow$} & \makecell{AUC $\uparrow$} & \makecell{SPD $\downarrow$} & \makecell{EOdds $\downarrow$} \\ 
 \hline
\multirow{9}{*}{Gender} 
& Baseline & 0.871$\pm$0.005 & 0.013$\pm$0.006 & 0.013$\pm$0.007 & 0.785$\pm$0.012 & 0.031$\pm$0.012 & 0.041$\pm$0.016\\
& FullFT-RW~\citep{zhang2022improving} & 0.865$\pm$0.009 & 0.014$\pm$0.006 & 0.014$\pm$0.008 & 0.789$\pm$0.011 & 0.032$\pm$0.013 & 0.041$\pm$0.015 \\
& FullFT-Reg~\citep{cherepanova2021technical} & 0.867$\pm$0.007 & 0.011$\pm$0.005 & 0.012$\pm$0.007 & 0.789$\pm$0.011 & 0.032$\pm$0.016 & 0.038$\pm$0.014 \\
& FullFT-FDR~\citep{le2023last} & 0.865$\pm$0.009 & 0.014$\pm$0.006 & 0.014$\pm$0.008 & 0.790$\pm$0.011 & 0.032$\pm$0.014 & 0.041$\pm$0.014 \\
& LLFT~\citep{mao2023last} & \underline{0.880}$\pm$0.006 & 0.014$\pm$0.005 & \underline{0.009}$\pm$0.004 & 0.786$\pm$0.008 & 0.033$\pm$0.009 & 0.048$\pm$0.012 \\
& DiffGda~\citep{marcinkevics2022debiasing}
&0.867$\pm$0.008 & 0.010$\pm$0.005 & 0.012$\pm$0.008 & 0.788$\pm$0.011 & 0.030$\pm$0.015 & 0.040$\pm$0.015 \\
& FairPrune~\citep{wu2022fairprune}
&0.860$\pm$0.008 & 0.019$\pm$0.008 & 0.012$\pm$0.004 & 0.781$\pm$0.012 & \underline{0.017}$\pm$0.010 & \underline{0.035}$\pm$0.015 \\
&DiffPrune~\citep{marcinkevics2022debiasing}&
0.855$\pm$0.007 & \underline{0.007}$\pm$0.004 & 0.009$\pm$0.005 & 0.773$\pm$0.022 & \textbf{0.014}$\pm$0.004 & 0.036$\pm$0.016\\
&
BMFT~\citep{xue2024BMFT}&
0.876$\pm$0.005 & \textbf{0.004}$\pm$0.004 & 0.011$\pm$0.004 & \textbf{0.797}$\pm$0.008 & 0.033$\pm$0.014 & 0.036$\pm$0.011\\
& SWiFT (Ours) & \textbf{0.880}$\pm$0.005 & 0.010$\pm$0.005 & \textbf{0.007}$\pm$0.004 & \underline{0.791}$\pm$0.011 & 0.040$\pm$0.016 & \textbf{0.035}$\pm$0.008  \\
\midrule
\multirow{2}{*}{} & \multirow{2}{*}{} & \multicolumn{3}{c}{Chexpert} & \multicolumn{3}{c}{NIH} \\ \cmidrule(lr){3-5} \cmidrule{6-8}
 &  & \makecell{AUC $\uparrow$} & \makecell{SPD $\downarrow$} & \makecell{EOdds $\downarrow$} & \makecell{AUC $\uparrow$} & \makecell{SPD $\downarrow$} & \makecell{EOdds $\downarrow$} \\ 
 \hline
\multirow{9}{*}{Age}& Baseline & \underline{0.871$\pm$0.005} & 0.109$\pm$0.023 & 0.051$\pm$0.011 & 0.785$\pm$0.012 & 0.063$\pm$0.024 & 0.078$\pm$0.018 \\
& FullFT-RW~\citep{zhang2022improving} & 0.871$\pm$0.006 & 0.106$\pm$0.023 & 0.050$\pm$0.009 & 0.786$\pm$0.011 & 0.061$\pm$0.023 & 0.071$\pm$0.016
\\
& FullFT-Reg~\citep{cherepanova2021technical}  & 0.865$\pm$0.007 & 0.098$\pm$0.021 & 0.045$\pm$0.006 & 0.790$\pm$0.011& \textbf{0.047}$\pm$0.020 & 0.077$\pm$0.011 \\
& FullFT-FDR~\citep{le2023last} & 0.869$\pm$0.006 & 0.098$\pm$0.021 & 0.044$\pm$0.005 & 0.788$\pm$0.011 & 0.050$\pm$0.020 & 0.071$\pm$0.013 \\
& LLFT~\citep{mao2023last} & 0.867$\pm$0.014 & 0.119$\pm$0.019 & 0.056$\pm$0.015 & 0.782$\pm$0.007 & 0.083$\pm$0.014 & 0.081$\pm$0.004 \\
& DiffGda~\citep{marcinkevics2022debiasing} & 0.862$\pm$0.009 & 0.091$\pm$0.027 & 0.044$\pm$0.006 & 0.789$\pm$0.011 & 0.050$\pm$0.026 & 0.075$\pm$0.012 \\
& FairPrune~\citep{wu2022fairprune} &0.837$\pm$0.052 & \underline{0.090}$\pm$0.029 & 0.044$\pm$0.006 & 0.776$\pm$0.010& 0.071$\pm$0.017 & 0.071$\pm$0.025 \\
& DiffPrune~\citep{marcinkevics2022debiasing} & 0.865$\pm$0.006 & 0.101$\pm$0.013 & 0.042$\pm$0.006 & 0.778$\pm$0.016 & 0.054$\pm$0.018 & \underline{0.063}$\pm$0.016 \\
&
BMFT~\citep{marcinkevics2022debiasing} & 0.870$\pm$0.004 & 0.108$\pm$0.009& \underline{0.038}$\pm$0.008 & \underline{0.791}$\pm$0.015 & 0.057$\pm$0.011 & 0.066$\pm$0.011 \\
& SWiFT (Ours)
& \textbf{0.873}$\pm$0.003 & \textbf{0.059}$\pm$0.019 & \textbf{0.025}$\pm$0.008 & \textbf{0.792}$\pm$0.011 & \underline{0.049}$\pm$0.017 & \textbf{0.062}$\pm$0.011\\
 \bottomrule
\end{tabular}
\end{adjustbox}
}
\end{table*}

\begin{table*}[ht]
\caption{Comparison of debiasing methods on gender and age attributes. DenseNet-121 model pre-trained on MIMIC
datasets.}\label{tab4}
\centering
{\begin{adjustbox}{width=1.0\textwidth}
\begin{tabular}{clllllll}
\toprule
 \multirow{2}{*}{Attr.} &  \multirow{2}{*}{Methods} & \multicolumn{3}{c}{Chexpert} & \multicolumn{3}{c}{NIH} \\ \cmidrule(lr){3-5} \cmidrule{6-8}
 &  & \makecell{AUC $\uparrow$} & \makecell{SPD $\downarrow$} & \makecell{EOdds $\downarrow$} & \makecell{AUC $\uparrow$} & \makecell{SPD $\downarrow$} & \makecell{EOdds $\downarrow$} \\ 
 \hline
\multirow{9}{*}{Gender} 
& Baseline & 0.880$\pm$0.006 & 0.012$\pm$0.011 & 0.010$\pm$0.005 & \underline{0.803}$\pm$0.005 & 0.030$\pm$0.008 & 0.027$\pm$0.007\\
& FullFT-RW~\citep{zhang2022improving} & 0.880$\pm$0.006 & 0.015$\pm$0.011 & 0.011$\pm$0.007 & 0.801$\pm$0.005 & 0.033$\pm$0.008 & 0.030$\pm$0.007 \\
& FullFT-Reg~\citep{cherepanova2021technical} & 0.879$\pm$0.004 & 0.010$\pm$0.007 & 0.011$\pm$0.006 & 0.801$\pm$0.005 & 0.034$\pm$0.010 & 0.027$\pm$0.008 \\
& FullFT-FDR~\citep{le2023last} & 0.878$\pm$0.005 & \textbf{0.008}$\pm$0.005 & 0.014$\pm$0.008 & 0.802$\pm$0.005 & 0.034$\pm$0.008 & 0.029$\pm$0.006 \\
& LLFT~\citep{mao2023last} & \underline{0.884}$\pm$0.004 & 0.019$\pm$0.009 & 0.011$\pm$0.006 & 0.795$\pm$0.006 & 0.033$\pm$0.009 & 0.029$\pm$0.009 \\
& DiffGda~\citep{marcinkevics2022debiasing}
&0.875$\pm$0.004 & 0.015$\pm$0.016 & 0.013$\pm$0.010 & 0.802$\pm$0.005 & 0.034$\pm$0.009 & 0.029$\pm$0.008 \\
& FairPrune~\citep{wu2022fairprune}
&0.715$\pm$0.113 & 0.018$\pm$0.019 & 0.019$\pm$0.020 & 0.796$\pm$0.005 & \textbf{0.024}$\pm$0.011 & \textbf{0.024}$\pm$0.009\\
&DiffPrune~\citep{marcinkevics2022debiasing}&
0.879$\pm$0.007 & 0.012$\pm$0.011 & \underline{0.009}$\pm$0.006 & 0.802$\pm$0.005 & \underline{0.029}$\pm$0.008 & \underline{0.026}$\pm$0.006\\
&
BMFT~\citep{xue2024BMFT}&
0.883$\pm$0.005 & 0.011$\pm$0.005 & 0.009$\pm$0.009 & \underline{0.805}$\pm$0.006 & 0.032$\pm$0.008 & 0.026$\pm$0.011\\
& SWiFT (Ours) & \textbf{0.887}$\pm$0.003 & \underline{0.009}$\pm$0.005 & \textbf{0.009}$\pm$0.003 &  \textbf{0.805}$\pm$0.004 & 0.039$\pm$0.004 & 0.027$\pm$0.003\\
\midrule
\multirow{2}{*}{} & \multirow{2}{*}{} & \multicolumn{3}{c}{Chexpert} & \multicolumn{3}{c}{NIH} \\ \cmidrule(lr){3-5} \cmidrule{6-8}
 &  & \makecell{AUC $\uparrow$} & \makecell{SPD $\downarrow$} & \makecell{EOdds $\downarrow$} & \makecell{AUC $\uparrow$} & \makecell{SPD $\uparrow$} & \makecell{EOdds $\downarrow$} \\ 
 \hline
\multirow{9}{*}{Age}& Baseline & \textbf{0.880}$\pm$0.006 & 0.120$\pm$0.011 & 0.048$\pm$0.011 & \underline{0.803}$\pm$0.005 & 0.071$\pm$0.009 & 0.065$\pm$0.013 \\
& FullFT-RW~\citep{zhang2022improving} & \underline{0.879}$\pm$0.004 & 0.124$\pm$0.008 & 0.052$\pm$0.008 & 0.799$\pm$0.004 & 0.073$\pm$0.006 & 0.068$\pm$0.015 \\
& FullFT-Reg~\citep{cherepanova2021technical}  & 0.875$\pm$0.005 & 0.115$\pm$0.008 & 0.044$\pm$0.009 & 0.799$\pm$0.005 & \underline{0.050}$\pm$0.010 & 0.065$\pm$0.024 \\
& FullFT-FDR~\citep{le2023last} & 0.877$\pm$0.004 & 0.116$\pm$0.010 & 0.045$\pm$0.010 & 0.799$\pm$0.006 & \textbf{0.048}$\pm$0.011 & 0.062$\pm$0.027 \\
& LLFT~\citep{mao2023last} & 0.869$\pm$0.007 & 0.130$\pm$0.006 & 0.061$\pm$0.006 & 0.801$\pm$0.004 & 0.080$\pm$0.010 & 0.061$\pm$0.009 \\
& DiffGda~\citep{marcinkevics2022debiasing} & 0.874$\pm$0.005 & 0.118$\pm$0.009 & 0.047$\pm$0.009 & 0.802$\pm$0.005 & 0.064$\pm$0.006 & 0.065$\pm$0.020 \\
& FairPrune~\citep{wu2022fairprune} &0.829$\pm$0.049 & 0.123$\pm$0.037 & 0.063$\pm$0.027 & 0.771$\pm$0.044 & 0.068$\pm$0.020 & 0.058$\pm$0.020 \\
& DiffPrune~\citep{marcinkevics2022debiasing} & 0.872$\pm$0.010 & \underline{0.111}$\pm$0.011 & 0.043$\pm$0.010 & 0.801$\pm$0.005 & 0.071$\pm$0.010 & \textbf{0.058}$\pm$0.014 \\
& BMFT~\citep{xue2024BMFT} & 0.877$\pm$0.008 & 0.116$\pm$0.008 & \underline{0.043}$\pm$0.006 & \underline{0.803}$\pm$0.005 & 0.070$\pm$0.013 & 0.060$\pm$0.008 \\
& SWiFT (Ours)  & 0.879$\pm$0.006 & \textbf{0.111}$\pm$0.011 & \textbf{0.042}$\pm$0.011 & \textbf{0.804}$\pm$0.002 & 0.061$\pm$0.009 & \underline{0.058}$\pm$0.019\\
 \bottomrule
\end{tabular}
\end{adjustbox}
}
\end{table*}
Table~\ref{tab1} through Table~\ref{tab4} report predictive accuracy and fairness scores achieved by all debiasing methods across various data modalities, sensitive attributes, and network architectures. Overall, our 
method demonstrates substantial improvements in fairness while maintaining high accuracy. For example, in debiasing skin tone using ResNet-50, SWiFT achieves an SPD and EOdds of 0.092 and 0.064, representing reductions of 36.1\% and 37.6\% compared to the Baseline model's 0.144 and 0.101, on Fitzpatrick 17k, respectively. Additionally, these values are 15.1\% and 27.3\% better than those achieved by the best SOTA method. The AUC of SWiFT improves upon other methods in most cases, and consistently outperforms the pre-trained Baseline model. Similar trends are observed in chest X-ray debiasing where SWiFT achieves improvements of 34.4\% in SPD and 34.2\% in EOdds compared to the best SOTA when debiasing the age attribute using ResNet-50.
Our preliminary method, BMFT, similarly achieves a desirable balance between debiasing and classification performance, with improved AUC and fairness metrics compared to other baselines. However, our proposed SWiFT consistently outperforms BMFT with a higher AUC and lower SPD and EOdds on most datasets, proving the effectiveness of our methodological modifications. Additionally, SWiFT eliminates the need for hyperparameter tuning of the hard-mask threshold, providing a more flexible and adaptive masking approach.

\textcolor{revisecolor}{In cases where SWiFT does not achieve the highest AUC or fairness scores, the leading alternate method often exhibits trade-offs, i.e.,~they show improved performance in one metric yet degraded performance elsewhere, sometimes performing even worse than the Baseline. For example, DiffPrune achieves the best SPD in debiasing gender with ResNet-50 backbone on the Atlas dataset, however, it demonstrates a $2.5\%$ decrease in AUC, 
with respect to  the Baseline. This illustrates a critical flaw in these alternate approaches: they often achieve fairness by `leveling down', a process that degrades the accuracy across all subgroups with a greater degradation occurring for the better performing groups. Such methods do not force the model to learn robust, unbiased representations but rather risk finding a simplistic solution where all groups perform equally poorly~\citep{zietlow2022leveling}. 
In contrast, our method consistently demonstrates improved AUC over the pre-trained model baseline in nearly all settings. This indicates that SWiFT can enhance fairness without sacrificing the model's discriminative capabilities.  
We further note one exception: on the CheXpert dataset, SWiFT with a DenseNet-121 backbone showed a slight decrease in AUC (0.001) compared to the Baseline when debiasing for age. We conjecture that this minor degradation is likely due to the DenseNet architecture. The densely-connected nature and large number of parameters may make it prone to overfitting and more difficult to optimize, particularly when learning new debiased features~\citep{yuan2019effective}. These factors lead to poor generalization, especially when fine-tuning on smaller datasets.
Despite this, SWiFT still outperformed other debiasing methods in this challenging scenario. The trade-off is minimal and limited to this specific architectural choice, underscoring the overall robustness of our approach.}


\noindent\textcolor{mycolor}{\textbf{Pruning is not always the best strategy} We observed that pruning-based methods (e.g., Diffprune, FairPrune) underperform fine-tuning-based methods in AUC on most test datasets (Tables~\ref{tab1}--\ref{tab4}), revealing that pruning often fails to preserve the discriminative capability of the pre-trained model. Furthermore, pruning methods exhibit high variance across different folds of cross-validation, indicating that their performance is neither stable nor generalizable across different environments.} 

\noindent\textcolor{mycolor}{\textbf{Fine-tuning utility.} 
\textcolor{revisecolor}{Our experiments analyzed the effectiveness of different fine-tuning strategies. First, the simplistic approach of fine-tuning on a balanced dataset (i.e., FullFT-RW) yields inconsistent results. While it demonstrates fairness gains on some datasets, it sometimes degrades fairness performance even below that of the original pre-trained model (i.e., Baseline). This indicates that when a model has already learned strong spurious correlations, re-weighting of the dataset, alone, is insufficient to unlearn them. The model may instead overfit to the previously learned biased representations, leading to poor generalization and fairness on OOD datasets.
In contrast, those integrating fairness constraints (i.e.~FulFT-Reg, FullFT-FDR and DiffGda) consistently achieve lower EOdds than those relying solely on cross-entropy loss (i.e.~FullFT-RW). This illustrates that explicitly incorporating the bias-related terms function into the loss is a more effective debiasing strategy than relying solely on data balancing.} Moreover, LLFT generally yields higher AUC and lower bias compared to full fine-tuning methods FullFT-RW, FullFT-Reg and FullFT-FDR, highlighting the importance of mask fine-tuning to avoid overfitting and maintain a better trade-off between classification performance and fairness. However, LLFT's poor AUC in certain scenarios indicate that fine-tuning only the last layer from scratch requires that core features are well captured and isolated from bias features. 
In contrast, our proposed method fine-tunes both the feature extractor and the last layer, which mitigates the bias from different components of the model and exhibits the best performance in terms of AUC, SPD and EOdds across most datasets.}

\noindent\textbf{Different bias, different difficulty.} Comparing Baseline results for gender and age attribute debiasing in Chexpert using ResNet50 (see Tables~\ref{tab3}), the pre-trained model demonstrates less inherent bias for gender when compared to age, as indicated by lower SPD and EOdds. Furthermore, Table~\ref{tab3} shows that SWiFT achieves better EOdds improvements for the age attribute c.f.~gender, e.g., EOdds is 45.9\% lower than Baseline for age while is 23.1\% for gender. This suggests that attributes with more inherent bias in the pre-trained model are easier for bias mitigation strategies to identify and rectify. Conversely, less inherent bias in the pre-trained model indicates that most core features are captured by the pre-trained model, which may limit the potential for AUC improvement.
\subsubsection{Qualitative Analysis}
\begin{figure}[t]
    \centering
    \includegraphics[width=\columnwidth,keepaspectratio]{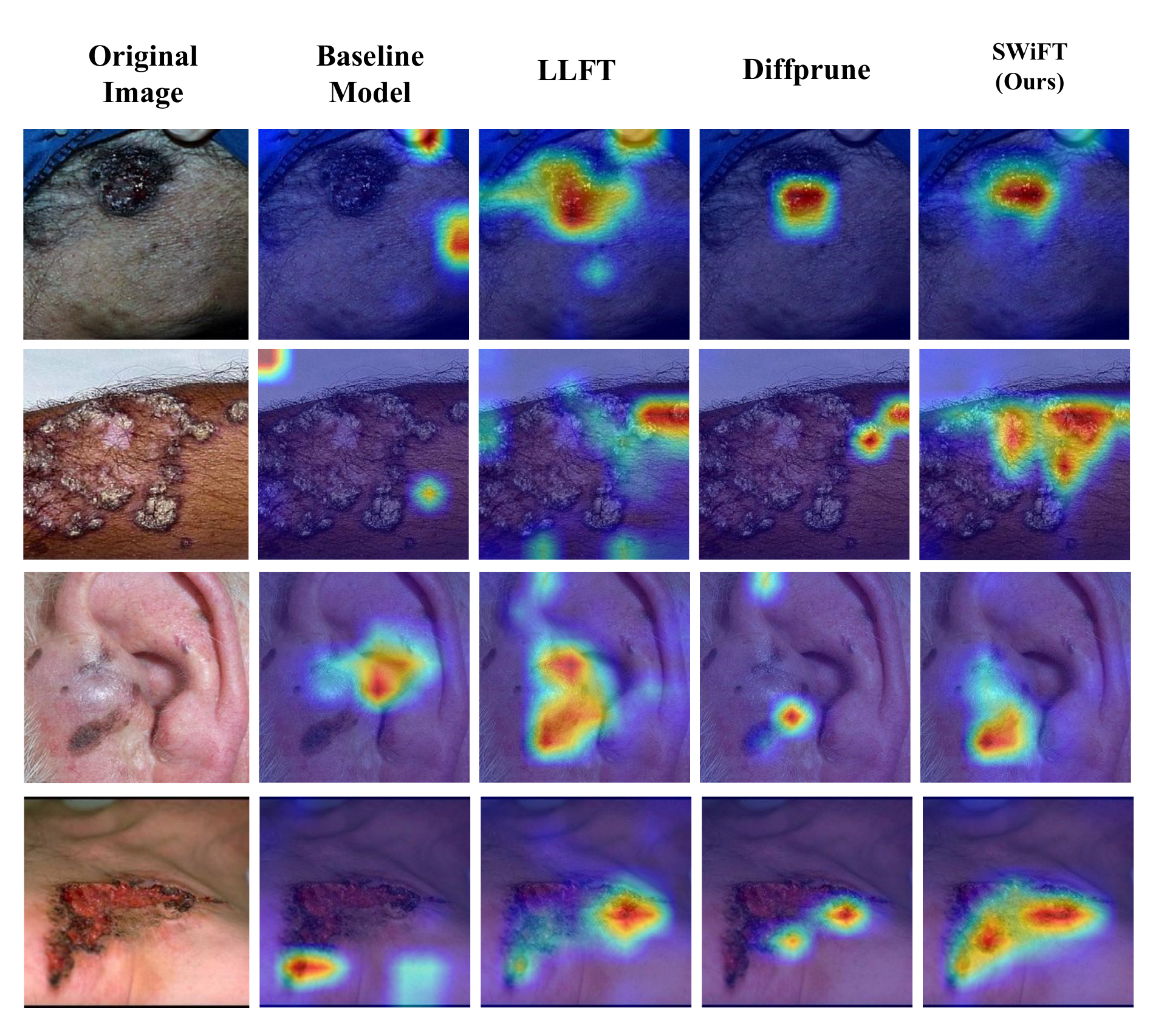}
    \caption{CAM visualization for four {Fitzpatrick-17k} test images. Original images (Original Image), CAM pre-trained model (Baseline Model), CAM fine-tuning based SOTA Method (LLFT~\citep{mao2023last}), CAM pruning-based SOTA Method (Diffprune~\citep{marcinkevics2022debiasing}), CAM post-debiasing model (SWiFT (Ours)). High, low activation indicated by red, blue respectively. 
    }
    \label{fig.Cam}
\end{figure}

\textcolor{mycolor}{Figure~\ref{fig.Cam} shows exemplary Class Activation Map (CAM)~\citep{zhou2016learning} visualizations from the skin classification task. We select images from both the benign and malignant class, as well as different skin attribute groups. The qualitative results in general follow the same behavior of the quantitative results described earlier: Our proposed method provides the best visual results, compared to the best fine-tuning-based method LLFT and the best pruning-based method Diffprune. Specifically, SWiFT successfully redirects attention away from unwanted features, i.e.,~skin tone, and towards core features, i.e.,~lesion areas. By constrast, DiffPrune demonstrates smaller activation regions and diminished focus on core features, which indicates that it
indiscriminately removes neurons responsible for \textit{both} bias and core features,  leading to loss of information necessary for accurate classification. Although LLFT outperforms the Baseline and DiffPrune with better visual lesion areas, it can still struggle to capture core features when the Baseline model has not learned them robustly.
\textcolor{revisecolor}{These observations provide clear evidence for the trade-offs detailed in our quantitative results. While prior SOTA methods tend to sacrifice the model's focus on core features to improve fairness, our method promotes better representation learning and improves overall performance.}}

\subsubsection{Extension to multiple attributes}
\label{sec:discussion:multi-attribute}
\textcolor{mycolor}{In the previous sections, we assumed that the sensitive attribute is binary. 
Towards exploring realistic settings, with more complex fairness-related issues, we further discuss the generalization of our method to multi-attribute scenarios. We consider a problem framing that allows us to reduce the multi-attribute problem to multiple instances of binary-attribute debiasing. 
It has been previously observed that solely optimizing for worst-group performance can effectively address distribution shifts and mitigate sub-group performance disparities~\citep{sagawa2019distributionally, liu2021just}. We therefore focus on optimizing the worst group via guidance from the best group. Specifically, we reduce the problem to a binary attribute task through selection of the best and worst groups, 
as defined by AUC, and use the samples from these two groups for optimization. We validate the multi-attribute generalization ability of our method on six skin tone groups and five age groups. As Table \ref{tab:multi} shows, our method is consistently effective in both overall AUC, the worst-group AUC, and the AUC gap between the best and the worst group. }
\begin{table}[ht]
\caption{Comparison of performance of non-binary attributes of skin tone for the skin lesion classification task and age for the chest X-ray classification task. ResNet-50 backbone.}\label{tab:multi}
\centering
{\begin{adjustbox}{width=\columnwidth}
\begin{tabular}{cllll}
\toprule
 \multirow{2}{*}{Attr.} &  \multirow{2}{*}{Methods} & \multicolumn{3}{c}{Fitzpatrick-17k}  \\ \cmidrule(lr){3-5} 
 &  & \makecell{Overall AUC $\uparrow$} & \makecell{Min. AUC $\uparrow$} & \makecell{Gap. AUC $\downarrow$} \\ 
 \hline
\multirow{2}{*}{Skin Tone} 
& Baseline  & 0.586\scriptsize$\pm$0.008 & 0.589$\pm$\scriptsize0.007 & 0.048$\pm$\scriptsize0.010 \\
& SWiFT (Ours) & \textbf{0.599}\scriptsize$\pm$0.025 & \textbf{0.596}\scriptsize$\pm$0.021 & \textbf{0.030}\scriptsize$\pm$0.005\\
\midrule
\multirow{2}{*}{} & \multirow{2}{*}{} & \multicolumn{3}{c}{Chexpert}\\ \cmidrule(lr){3-5}
 &  &  \makecell{Overall AUC $\uparrow$} & \makecell{Min. AUC $\uparrow$} & \makecell{Gap. AUC $\downarrow$}\\ 
 \hline
 \multirow{2}{*}{Age} 
 & Baseline  & 0.871\scriptsize$\pm$0.005 & 0.813\scriptsize$\pm$0.005 & 0.072\scriptsize$\pm$0.007\\
& SWiFT (Ours) &\textbf{0.880}\scriptsize$\pm$0.001 & \textbf{0.825}\scriptsize$\pm$0.002 & \textbf{0.068}\scriptsize$\pm$0.004\\
 \bottomrule
\end{tabular}
\end{adjustbox}
}
\end{table}

\subsection{Ablation Study}
\label{sec:discussion:ablate}
We conduct ablative studies on both skin lesion and chest X-ray classification tasks to evaluate the effectiveness of SWiFT in key aspects pertaining to soft-mask utility, efficacy of our fine-tuning components, and the adaptability to external dataset sizes. To this end, we investigate: 1) soft-mask performance compared to the random-mask and hard-mask strategies, and 2) how each step of our fine-tuning process individually impacts fairness and discriminative performance, and 3) performance sensitivity to different sizes of the external dataset.
\subsubsection{Sensitivity to masking strategy}
\label{sec:discussion:ablate:mask} 
We use a soft-mask to preserve discriminative capability while performing debiasing. To evaluate efficacy, we performed a sensitivity study comparing our soft-mask with both a random and a hard-mask. A random soft-mask assigns values by randomly sampling from the $[0,1]$ range. A hard-mask converts the importance scores into binary values based on a predefined threshold. Parameters with importance scores above the threshold are adjusted, while others remain fixed. We search for the optimal fine-tuning rates in $\{0.1, 0.3, 0.5, 0.7, 0.9\}$. The results of the ablation experiments are shown in Table~\ref{tab5}. 
Both soft- and hard-mask strategies outperform random-mask, illustrating the effectiveness of our mask generation process.
Moreover, hard-mask performs worse than soft-mask with higher SPD, EOdds and much lower AUC. This shows the difficulty of finding optimal binary mask thresholds. 
Soft-masking 
proves to be more flexible and effective in achieving improved fairness-prediction trade-offs. 
\begin{table}[ht]
\caption{Comparison of random, hard-masks and our proposed mask generation method on skin and gender attributes for the skin lesion classification task. ResNet-50 backbone.}\label{tab5}
\centering
{\begin{adjustbox}{width=\columnwidth}
\begin{tabular}{cllll}
\toprule
 \multirow{2}{*}{Attr.} &  \multirow{2}{*}{Methods} & \multicolumn{3}{c}{Fitzpatrick-17k}  \\ \cmidrule(lr){3-5} 
 &  & \makecell{AUC $\uparrow$} & \makecell{SPD $\downarrow$} & \makecell{EOdds $\downarrow$} \\ 
 \hline
\multirow{3}{*}{Skin Tone} 
& Random  & 0.615\scriptsize$\pm$0.012 & 0.119$\pm$\scriptsize0.016 & 0.128$\pm$\scriptsize0.011\\
 & hard-mask  & 0.622\scriptsize$\pm$0.002 & 0.099$\pm$\scriptsize0.017 & 0.102$\pm$\scriptsize0.017 \\
 & SWiFT (Ours) & \textbf{0.668}$\pm$\scriptsize0.013 & \textbf{0.092}$\pm$\scriptsize0.028 & \textbf{0.096}$\pm$\scriptsize0.026 \\
\midrule
\multirow{2}{*}{} & \multirow{2}{*}{} & \multicolumn{3}{c}{Atlas}\\ \cmidrule(lr){3-5}
 &  & \makecell{AUC $\uparrow$} & \makecell{SPD $\downarrow$} & \makecell{EOdds $\downarrow$}\\ 
 \hline
 \multirow{3}{*}{Gender} 
 & Random & 0.787\scriptsize$\pm$0.004 & 0.017\scriptsize$\pm$0.009 & 0.017\scriptsize$\pm$0.008\\
 & {hard-mask}   & 0.788\scriptsize$\pm$0.004 & 0.023\scriptsize$\pm$0.013 & 0.023\scriptsize$\pm$0.006 \\
 & SWiFT (Ours) &\textbf{0.789}\scriptsize$\pm$0.005 & \textbf{0.012}\scriptsize$\pm$0.007 & \textbf{0.010}\scriptsize$\pm$0.006\\
 \bottomrule
\end{tabular}
\end{adjustbox}
}
\end{table}

\subsubsection{Sensitivity to mask normalization strategy}
\label{sec:discussion:ablate:normal} 
\begin{table}[ht]
\caption{Comparison of random, hard-masks and our proposed mask generation method on skin and gender attributes for the skin lesion classification task. ResNet-50 backbone.}\label{tab:norm}
\centering
{\begin{adjustbox}{width=\columnwidth}
\begin{tabular}{cllll}
\toprule
 \multirow{2}{*}{Attr.} &  \multirow{2}{*}{Methods} & \multicolumn{3}{c}{Fitzpatrick-17k}  \\ \cmidrule(lr){3-5} 
 &  & \makecell{AUC $\uparrow$} & \makecell{SPD $\downarrow$} & \makecell{EOdds $\downarrow$} \\ 
 \hline
\multirow{2}{*}{Skin Tone} 
& z-score  & 0.655\scriptsize$\pm$0.012 & 0.109$\pm$\scriptsize0.011 & 0.105$\pm$\scriptsize0.009\\
 & SWiFT (Min--Max) & \textbf{0.668}$\pm$\scriptsize0.013 & \textbf{0.092}$\pm$\scriptsize0.028 & \textbf{0.096}$\pm$\scriptsize0.026 \\
\midrule
\multirow{2}{*}{} & \multirow{2}{*}{} & \multicolumn{3}{c}{Chexpert}\\ \cmidrule(lr){3-5}
 &  & \makecell{AUC $\uparrow$} & \makecell{SPD $\downarrow$} & \makecell{EOdds $\downarrow$}\\ 
 \hline
 \multirow{2}{*}{Age} 
 & z-score & \textbf{0.880}\scriptsize$\pm$0.005 & 0.087\scriptsize$\pm$0.028 & 0.040\scriptsize$\pm$0.007\\
 & SWiFT (Min--Max) & 0.873\scriptsize$\pm$0.003 & \textbf{0.059}\scriptsize$\pm$0.019 & \textbf{0.025}\scriptsize$\pm$0.008 \\
 \bottomrule
\end{tabular}
\end{adjustbox}
}
\end{table} 

\textcolor{revisecolor}{For our soft-mask construction, we employ Min-Max normalization over the commonly used z-score method. As shown in Table~\ref{tab:norm}, Min--Max normalization yields superior performance in both AUC and fairness metrics. 
We attribute this improvement to our mask's objective: determining the relative ratio between parameter importance for bias and for prediction. Min-Max normalization preserves the shape and relative relationships of the original parameter importance distribution~\citep{singh2020investigating}. In contrast, z-score normalization, which re-centers and rescales the data, can distort these crucial relationships. Recognizing that Min-Max normalization is sensitive to outliers, we will explore alternative strategies, such as percentile-based normalization, in future work.}

\subsubsection{Effectiveness of two-step combination}
\label{sec:discussion:ablate:twostep} 
SWiFT consists of two core steps that contribute to its performance: mask fine-tuning of the feature extractor and partial re-initialization followed by full classification head fine-tuning. \textcolor{revisecolor}{To assess the impact of each step, we conducted an ablation study where we compare performance of (i) feature extractor fine-tuning only, (ii) classification head re-initialization and fine-tuning only and (iii) the full pipeline.} The results, shown in Table \ref{tab: fine-tuning}, indicate a significant performance decline when either step is removed. However, each individual step still outperforms the Baseline, supporting our hypothesis that bias originates from both feature extraction and feature combination. The integration of both steps effectively eliminates bias from these two sources, resulting in superior performance in terms of both prediction accuracy and fairness.
\begin{table}[ht]
\caption{Comparison of performance when ablating fine-tuning components for both skin tone and gender attributes under the skin lesion classification task. ResNet-50 backbone.}\label{tab: fine-tuning}
\centering
{\begin{adjustbox}{width=\columnwidth}
\begin{tabular}{cllll}
\toprule
 \multirow{2}{*}{Attr.} &  \multirow{2}{*}{Methods} & \multicolumn{3}{c}{Fitzpatrick-17k}  \\ \cmidrule(lr){3-5} 
 &  & \makecell{AUC $\uparrow$} & \makecell{SPD $\downarrow$} & \makecell{EOdds $\downarrow$} \\ 
 \hline
\multirow{3}{*}{Skin Tone} 
& {SWiFT (w/o 2nd step)}  & 0.605\scriptsize$\pm$0.009 & 0.125$\pm$\scriptsize0.011 & 0.125$\pm$\scriptsize0.010 \\
& {SWiFT (w/o 1st step)}   & 0.586\scriptsize$\pm$0.005 & 0.146$\pm$\scriptsize0.009 & 0.146$\pm$\scriptsize0.007 \\
& SWiFT (Ours) & \textbf{0.668}$\pm$\scriptsize0.013 & \textbf{0.092}$\pm$\scriptsize0.028 & \textbf{0.096}$\pm$\scriptsize0.026\\
\midrule
\multirow{2}{*}{} & \multirow{2}{*}{} & \multicolumn{3}{c}{Atlas}\\ \cmidrule(lr){3-5}
 &  & \makecell{AUC $\uparrow$} & \makecell{SPD $\downarrow$} & \makecell{EOdds $\downarrow$}\\ 
 \hline
 \multirow{3}{*}{Gender} 
 & {SWiFT (w/o 2nd step)}  & 0.787\scriptsize$\pm$0.007 & 0.025\scriptsize$\pm$0.009 & 0.021\scriptsize$\pm$0.015\\
& {SWiFT (w/o 1st step)}   & 0.788\scriptsize$\pm$0.005 & 0.022\scriptsize$\pm$0.007 & 0.019\scriptsize$\pm$0.014 \\
& SWiFT (Ours) &\textbf{0.789}\scriptsize$\pm$0.005 & \textbf{0.012}\scriptsize$\pm$0.007 & \textbf{0.010}\scriptsize$\pm$0.006\\
 \bottomrule
\end{tabular}
\end{adjustbox}
}
\end{table}


\begin{figure*}[!htbp]
    \centering
    \begin{subfigure}[b]{0.45\textwidth}
        \centering
        \includegraphics[width=0.9\textwidth]{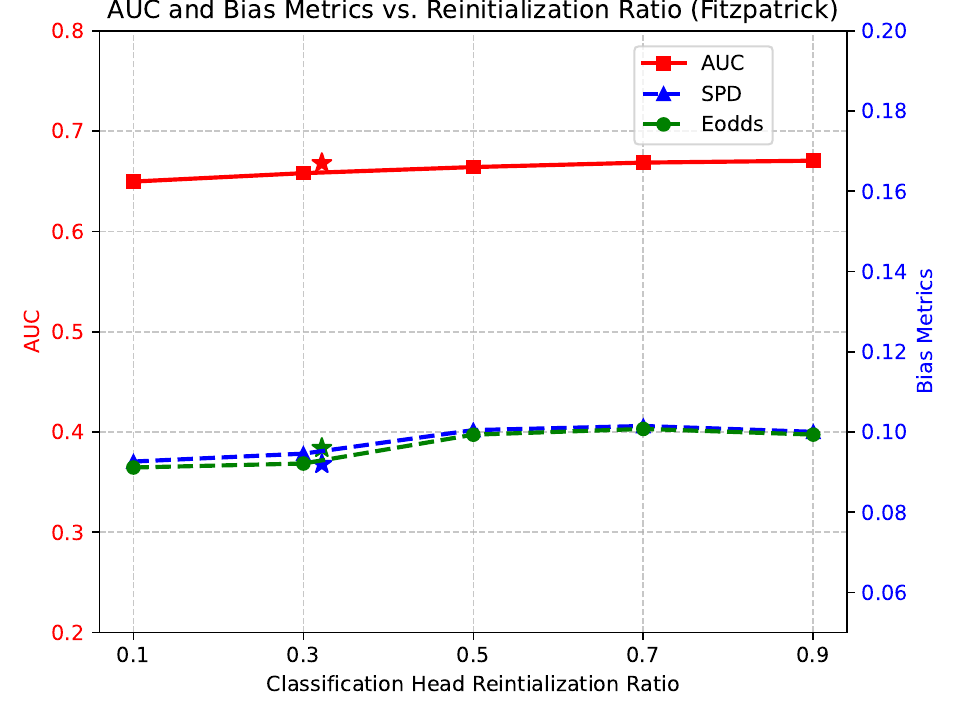}
        \caption{}
        \label{fig:ReinitRatio_Fitz}
    \end{subfigure}
    \begin{subfigure}[b]{0.45\textwidth}
        \centering
        \includegraphics[width=0.9\textwidth]{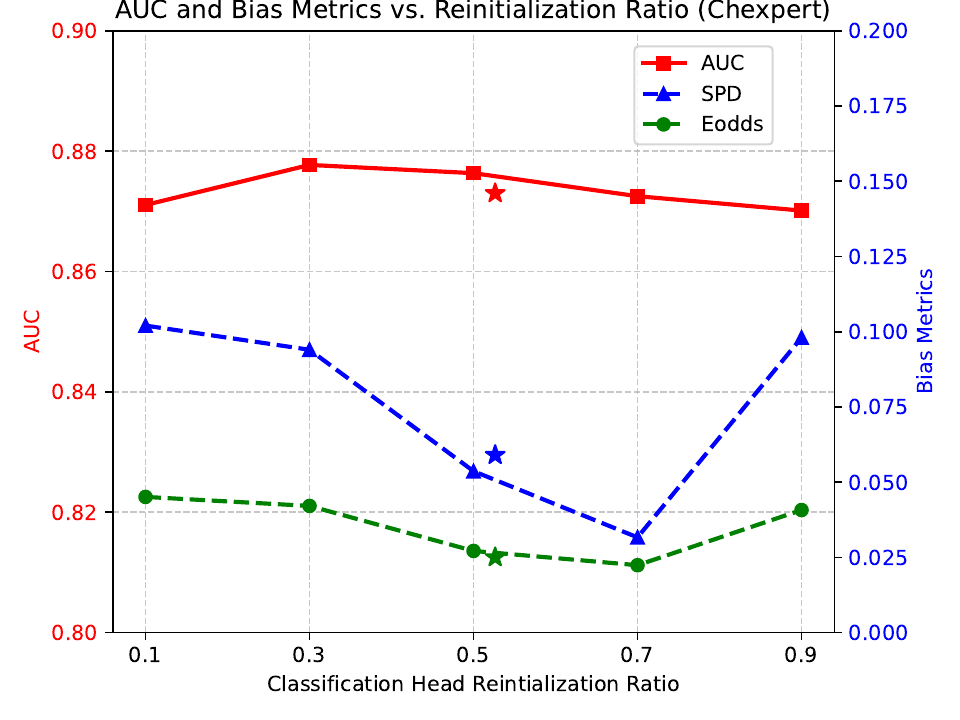}
        \caption{}
        \label{fig:ReinitRatio_Chex}
    \end{subfigure}
    
    \caption{\textcolor{mycolor}{Sensitivity analysis for the re-initialization threshold $\gamma$. AUC and fairness performance is plotted against the resulting proportion of re-initialized parameters in the classification head, under a ResNet-50 backbone. The x-axis represents the percentage of parameters re-initialized based on having the highest mask scores.  (a) Results for skin tone debiasing on the Fitzpatrick-17k dataset. (b) Results for age debiasing on the CheXpert dataset. The star ($\star$) indicates the performance of our proposed method, where the re-initialization proportion is automatically determined by the mean value of the soft-mask calculated on the classification head.}}
    \label{fig:reinitratio_comparison}
\end{figure*}

\subsubsection{Re-initialization effectiveness}
\label{sec:discussion:ablate:reinit} 
\textcolor{mycolor}{To evaluate the effectiveness of our partial re-initialization strategy, we compare SWiFT with two re-initialization variants, i.e.,~no classification head re-initialization, and full classification head re-initialization. The results are shown in Table ~\ref{tab: reinit}. As we can see, both full and partial re-initialization achieve higher accuracy and fairness than the no re-initialization approach, indicating that parameter re-initialization can effectively remove bias and recombine core predictive features. However, partial re-initialization achieves even better results, particularly in terms of AUC. This finding suggests that resetting all parameters in the classification head may lead to catastrophic forgetting of previously learned representations, which complicates the optimization process and ultimately results in suboptimal performance. In contrast, partial re-initialization preserves parameters, important for core features, thus providing favorable 
seeding for fast and effective fine-tuning. }

\begin{table}[ht]
\caption{\textcolor{mycolor}{Comparison of performance of partial re-initialization vs full re-initialization strategies of skin tone for the skin lesion classification task and age for the chest X-ray classification task. ResNet-50 backbone.}}\label{tab: reinit}
\centering
{\begin{adjustbox}{width=\columnwidth}
\begin{tabular}{cllll}
\toprule
 \multirow{2}{*}{Attr.} &  \multirow{2}{*}{Methods} & \multicolumn{3}{c}{Fitzpatrick-17k}  \\ \cmidrule(lr){3-5} 
 &  & \makecell{AUC $\uparrow$} & \makecell{SPD $\downarrow$} & \makecell{EOdds $\downarrow$} \\ 
 \hline
\multirow{3}{*}{Skin Tone} 
& No Reinit  & 0.608\scriptsize$\pm$0.019 & 0.112$\pm$\scriptsize0.019 & 0.103$\pm$\scriptsize0.019 \\
& Full Reinit  & 0.619\scriptsize$\pm$0.016 & 0.109$\pm$\scriptsize0.024 & 0.105$\pm$\scriptsize0.015 \\
& SWiFT (Partial) &  \textbf{0.668}$\pm$\scriptsize0.013 & \textbf{0.092}$\pm$\scriptsize0.028 & \textbf{0.096}$\pm$\scriptsize0.026\\
\midrule
\multirow{2}{*}{} & \multirow{2}{*}{} & \multicolumn{3}{c}{Chexpert}\\ \cmidrule(lr){3-5}
 &  &  \makecell{AUC $\uparrow$} & \makecell{SPD $\downarrow$} & \makecell{EOdds $\downarrow$}\\ 
 \hline
 \multirow{2}{*}{Age} 
 & No Reinit  &0.873\scriptsize$\pm$0.004 & 0.104\scriptsize$\pm$0.022 & 0.044\scriptsize$\pm$0.008 \\
& Full Reinit &0.876\scriptsize$\pm$0.004 & 0.106\scriptsize$\pm$0.017 & 0.043\scriptsize$\pm$0.007\\
& SWiFT (Partial)
& \textbf{0.873}\scriptsize$\pm$0.003 & \textbf{0.059}\scriptsize$\pm$0.019 & \textbf{0.025}\scriptsize$\pm$0.008 \\
 \bottomrule
\end{tabular}
\end{adjustbox}
}
\end{table}

\subsubsection{Ablation on the Re-initialization Threshold}
\label{sec:discussion:ablate:reinit-thresh} 
\textcolor{revisecolor}{We investigate the sensitivity of varying our classification head re-initialization threshold, $\gamma$. Our analysis compares our proposed heuristic (i.e., setting $\gamma$ to the mean of the mask) against a range of thresholds defined by quantiles of the mask distribution. This corresponds to re-initializing a varying proportion (from 10\% to 90\%) of classification head parameters with the highest soft-mask values. As shown in Figure~\ref{fig:reinitratio_comparison}, the optimal re-initialization ratio is task-dependent and varies across the Fitzpatrick-17k and CheXpert datasets. Notably, our proposed method of setting $\gamma$ to the mean value of the soft-mask calculated on the classification head parameters consistently yields near-optimal prediction and fairness performance without requiring an exhaustive parameter search. This demonstrates that using the mean mask value is a sufficiently effective choice.}

\begin{figure*}[ht]
    \centering
    \begin{subfigure}[b]{0.45\textwidth}
        \centering
        \includegraphics[width=0.9\textwidth]{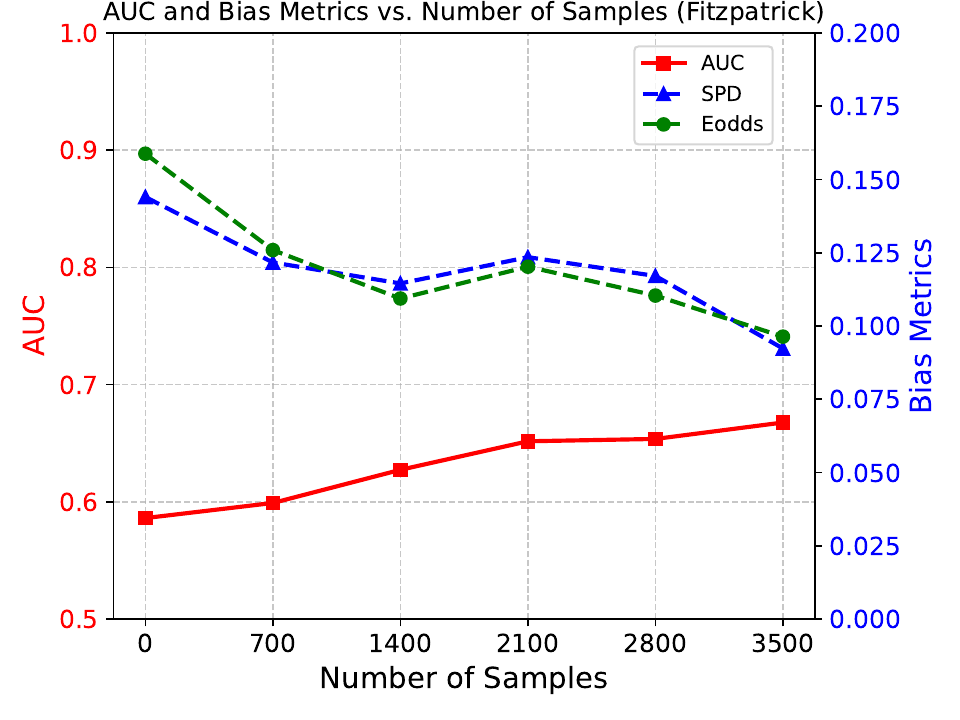}
        \caption{}
        \label{fig:NumOfSample_Fitz}
    \end{subfigure}
    \begin{subfigure}[b]{0.45\textwidth}
        \centering
        \includegraphics[width=0.9\textwidth]{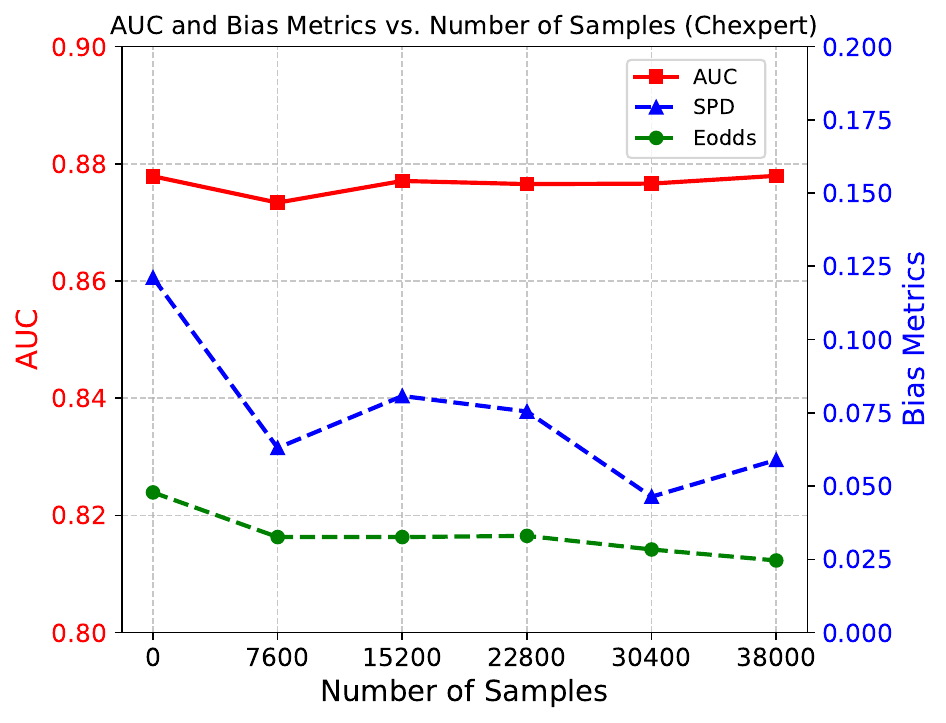}
        \caption{}
        \label{fig:NumOfSample_Chex}
    \end{subfigure}
    
    \caption{\textcolor{mycolor}{Comparison of performance of different size of the external dataset of skin tone for the skin lesion classification task and age for the chest X-ray classification task. ResNet-50 backbone. (a) is on the Fitzpatrick-17k dataset for skin tone debiasing, (b) is on the Chexpert dataset for age debiasing. }}
    \label{fig:numofsample_comparison}
\end{figure*}

\subsubsection{Ablation on the Number of Samples}
\label{sec:discussion:ablate:samplesize} 
\textcolor{mycolor}{Figure~\ref{fig:numofsample_comparison} demonstrates the sensitivity of SWiFT performance w.r.t.~number of samples on the external dataset. We conducted experiments using $\{0, 0.2, 0.4, 0.6, 0.8, 1\}$ proportions of the full external dataset, where the 0- sample condition corresponds to the pre-trained Baseline model. As the number of samples increases, the AUC improves while bias decreases. For skin tone, the AUC and bias reach stable after the dataset exceeds 0.4 of its total size (i.e., 1400 samples). Notably, even with a very small external dataset (i.e., $20\%$ of the dataset with 700 samples), SWiFT outperforms the Baseline in both accuracy and fairness. A similar trend emerges in age debiasing on Chexpert, where only $20\%$ of the validation dataset provides a significant improvement in fairness with minimal impact on AUC. The relatively modest increases in AUC across different sample sizes on CheXpert may stem from the already high baseline performance of the pre-trained model, leaving limited room for further improvement. Overall, these findings suggest that SWiFT can operate effectively with relatively small external datasets, highlighting its ease of construction and promising potential for practical advantages in real-world clinical deployments.}

\begin{figure}[!ht]
    \centering
    \includegraphics[width=\linewidth]{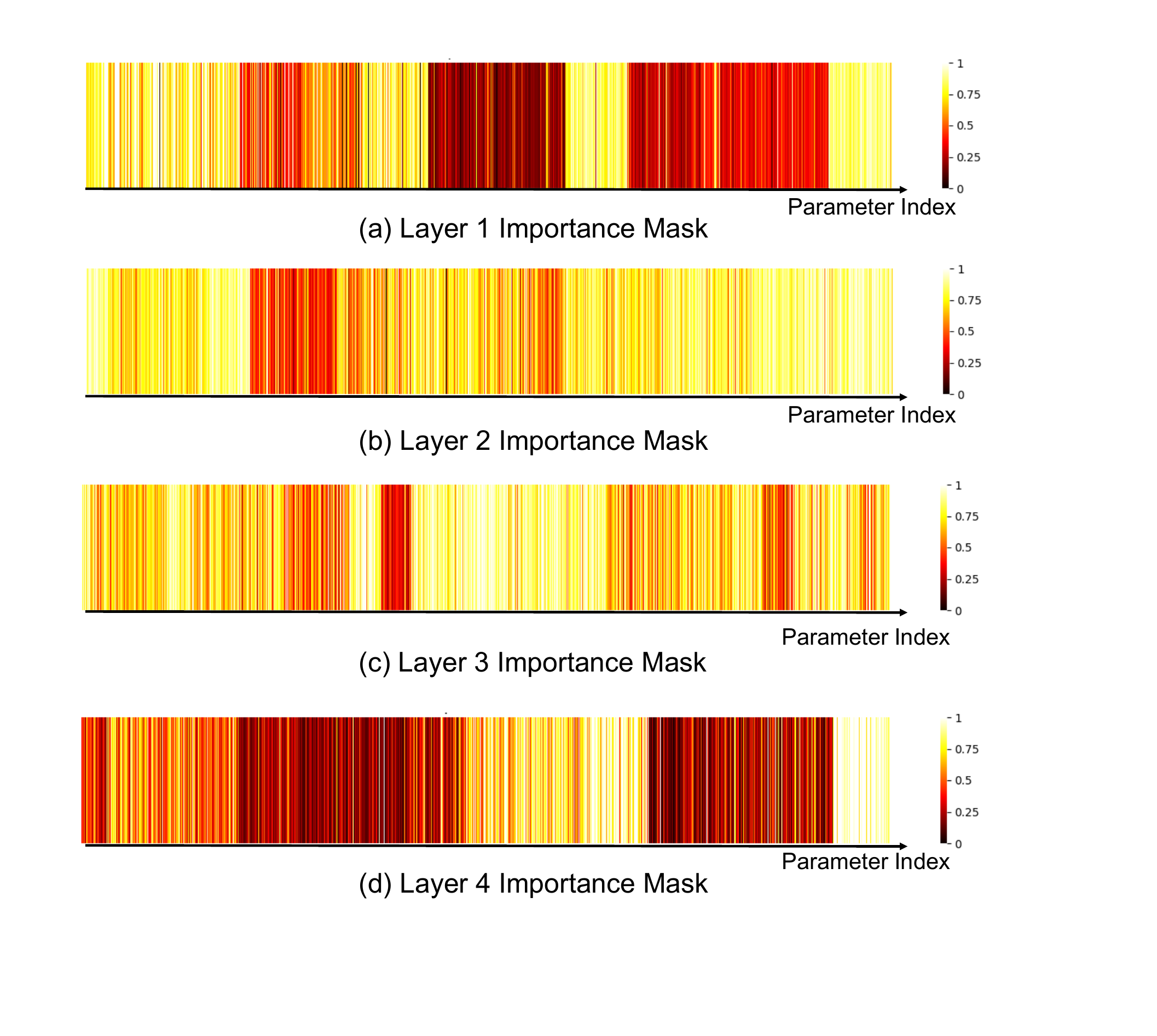}
    \caption{\textcolor{revisecolor}{An illustration of the soft-mask in each Layer (Residual block group/ConvN\_x) of the ResNet-50 on ISIC dataset for skin tone debiasing. The x-axis is the index of parameters in each layer of the ResNet-50. The color indicates the corresponding mask value for each parameter.}}
    \label{fig.layermaskisic}
\end{figure}

\begin{figure}[!ht]
    \centering
    \includegraphics[width=\linewidth]{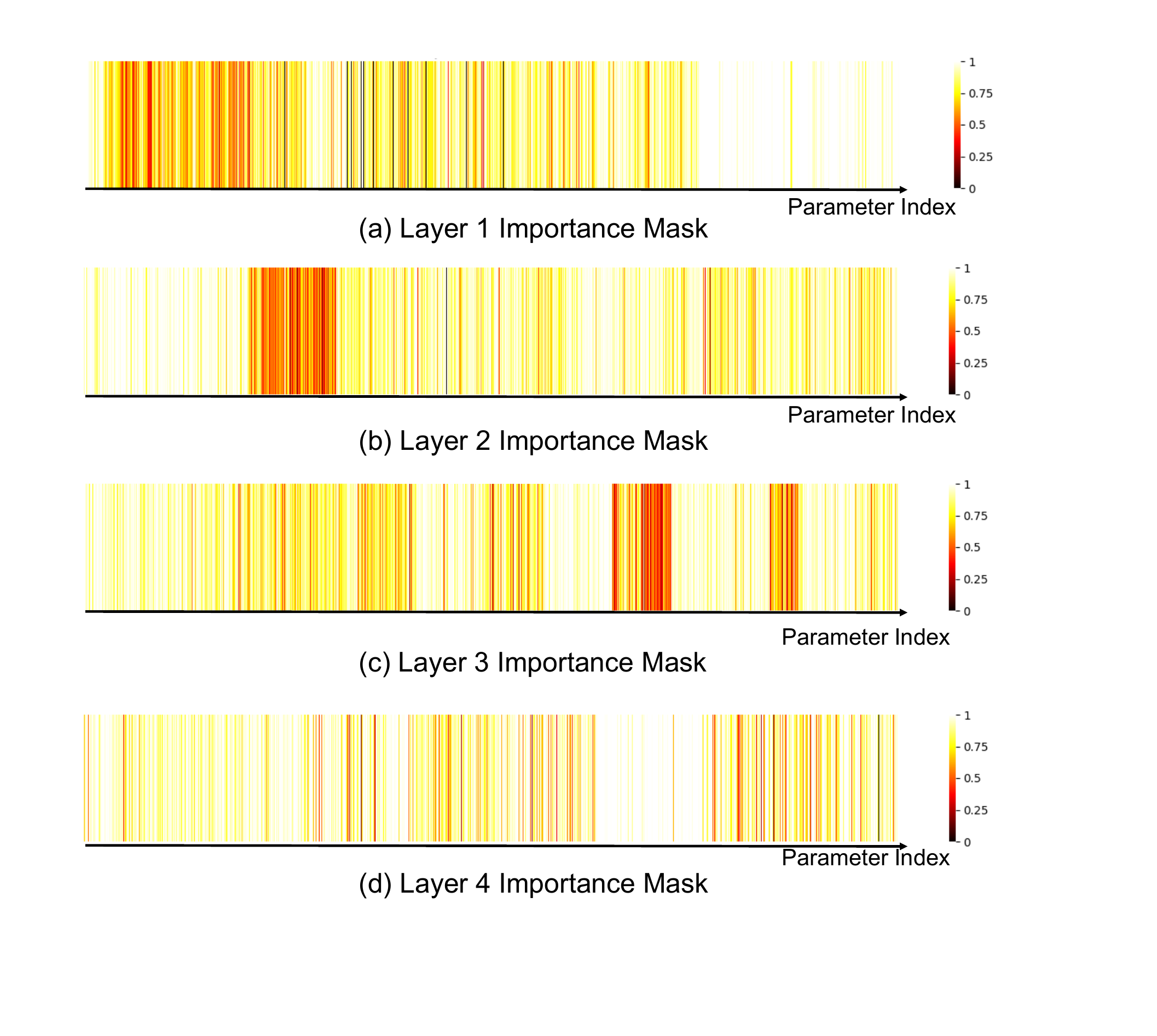}
    \caption{\textcolor{revisecolor}{An illustration of the soft-mask in each Layer (Residual block group/ConvN\_x) of the ResNet-50 on MIMIC dataset for age debiasing. The x-axis is the index of parameters in each layer of the ResNet-50. The color indicates the corresponding mask value for each parameter.}}
    \label{fig.layermaskchex}
\end{figure}
\subsection{Soft-mask Visualizations and Interpretations}
\textcolor{revisecolor}{We analyze the distribution of our soft-mask across the layers (residual block group) of a ResNet-50 model to understand how our method allocates parameter updates to address bias and improve core feature learning. Our findings reveal that our mask adaptively identifies different sources of bias depending on the task and model's initial state.}

\textcolor{revisecolor}{For the skin tone debiasing task on the ISIC dataset (Figure.~\ref{fig.layermaskisic}), the highest mask values are concentrated in the early-to-middle layers (e.g., Layer 2 and Layer 3). This suggests that the updates address the low-level features associated with skin tone bias. Moreover, as the pre-trained has a suboptimal initial AUC, updating shallow layers also improves core representations for better prediction accuracy. In contrast, for the age debiasing task on the MIMIC dataset (Figure.~\ref{fig.layermaskchex}), where the model already has a high initial AUC, the mask correctly focuses updates on the deeper layers (e.g., Layer 4). Since the model's core representations are well-learned, our mask targets the high-level semantic features encoded in deeper layers, which are more likely to be entangled with the abstract concept of age.} 

\textcolor{revisecolor}{
\section{Limitations and Future Work}
\label{sec:diss}
Our study has several limitations. First, our experiments were conducted primarily on binary classification tasks with binary sensitive attributes. Although our ablation study provides initial evidence that our method can extend to multi-attribute settings, a comprehensive evaluation on more complex multi-class classification problems is a key next step. However, the core mechanism of our method—disentangling parameter updates based on their importance to task performance versus fairness — is conceptually general. This framework is not intrinsically tied to any specific accuracy or fairness function; therefore, we believe our method can be extended to deal with more diverse prediction or bias settings. Second, our current implementation uses a fixed soft-mask, which is computed once before fine-tuning. This design is based on the assumption that, given the small number of fine-tuning epochs, the relative importance of parameters remains largely stable. This avoids the computational overhead of repeatedly recomputing the mask. However, exploring an adaptive masking strategy, where the mask is updated dynamically, remains a promising direction for future work, especially for applications require longer fine-tuning. Third, while setting the classification head re-initialization threshold to the mean mask value proves to be an effective strategy, future work could explore methods for learning this threshold adaptively. An adaptive threshold may provide a more fine-grained and performant approach to re-initialization. }

\section{Conclusion}
\label{sec:conc}
Our study advances bias mitigation in discriminative models trained on dermatological and chest X-ray data. Our method distinguishes core features from biased features, towards enhancing fairness without sacrificing classification performance. Our two-step fine-tuning approach reduces bias under small epoch counts (${\sim}10\%$ of original training compute), while remaining agnostic to the choice of model architecture. 
We thus present an efficient solution, applicable in resource limited scenarios. The challenge of obtaining diverse datasets with comprehensive metadata and sensitive attributes remains a limitation. 
However, unlike other methods, our approach requires only a small dataset for debiasing, partially alleviating this factor. 







\acks{The work of Junyu Yan was supported in part by the Advanced Care Research Center by the Ph.D. studentship. Sotirios A. Tsaftaris acknowledges support from the Royal Academy of Engineering and the Research Chairs and Senior Research Fellowships scheme (grant RCSRF1819\textbackslash8\textbackslash 25), and the UK Engineering and Physical Sciences Research Council (EPSRC) support via grant EP/X017680/1, and the UKRI AI programme, for the Causality in Healthcare AI Hub (CHAI, grant EP/Y028856/1).}

%
\ethics{The work follows appropriate ethical standards in conducting research and writing the manuscript, following all applicable laws and regulations regarding treatment of animals or human subjects.}

\coi{We declare we don’t have conflicts of interest.}

\data{All data used in the experiments is publicly available. The ISIC dataset, Fitzpatrick-17k dataset, and Atlas dataset can be obtained based on instructions
at https://github.com/pbevan1/Skin-Deep-Unlearning. The provided codebase contains implementation for all additional data pre-processing that was done. The DDI dataset can be accessed at https://ddi-dataset.github.io/. The PAD-UFES-20 dataset can be accessed at https://data.mendeley.com/datasets/zr7vgbcyr2/1. The MIMIC dataset can be downloaded from https://physionet.org/content/mimic-cxr/2.1.0/. The Chexpert dataset can be accessed at https://stanfordmlgroup.github.io/competitions/chexpert/. The NIH dataset can be accessed at https://www.kaggle.com/datasets/nih-chest-xrays/data}

\bibliography{melba-sample.bib}






\end{document}